\documentclass{article}

\usepackage[english]{babel}

\usepackage[letterpaper,top=2cm,bottom=2cm,left=3cm,right=3cm,marginparwidth=1.75cm]{geometry}

\usepackage{placeins}
\usepackage{amsmath}
\usepackage{authblk}
\usepackage{natbib}
\usepackage{graphicx}
\usepackage{tabularx}
\usepackage{makecell}
\usepackage{subcaption}
\usepackage{appendix}
\usepackage{multirow} 
\usepackage[colorlinks=true, allcolors=blue]{hyperref}
\usepackage{threeparttable}
\usepackage{booktabs}
\title{Leveraging Sidewalk Robots for Walkability-Related Analyses}

\author[a,c]{Xing Tong}
\author[a,c]{Michele D. Simoni\thanks{Corresponding author. Email: micheles@kth.se}}
\author[b,c]{Kaj Munhoz Arfvidsson}
\author[b,c]{Jonas Mårtensson}

\affil[a]{Division of Transport and Systems Analysis, KTH Royal Institute of Technology, Teknikringen 10A, Stockholm 10044, Sweden}
\affil[b]{Division of Decision and Control Systems, KTH Royal Institute of Technology, Malvinas väg 10, Stockholm 10044, Sweden}
\affil[c]{Integrated Transport Research Lab, KTH Royal Institute of Technology, Drottning Kristinas Väg 40, Stockholm 10044, Sweden}

\begin{document}

\maketitle

\begin{abstract}
Walkability is a key component of sustainable urban development. In walkability studies, collecting detailed pedestrian infrastructure data remains challenging due to the high costs and limited scalability of traditional methods. Sidewalk delivery robots, increasingly deployed in urban environments, offer a promising solution to these limitations. This paper explores how these robots can serve as mobile data collection platforms, capturing sidewalk-level features related to walkability in a scalable, automated, and real-time manner. A sensor-equipped robot was deployed on a sidewalk network at KTH in Stockholm, completing 101 trips covering 900 segment records. From the collected data, different typologies of features are derived, including robot trip characteristics (e.g., speed, duration), sidewalk conditions (e.g., width, surface unevenness), and sidewalk utilization (e.g., pedestrian density). Their walkability-related implications were investigated with a series of analyses. The results demonstrate that pedestrian movement patterns are strongly influenced by sidewalk characteristics, with higher density, reduced width, and surface irregularity associated with slower and more variable trajectories. Notably, robot speed closely mirrors pedestrian behavior, highlighting its potential as a proxy for assessing pedestrian dynamics. The proposed framework enables continuous monitoring of sidewalk conditions and pedestrian behavior, contributing to the development of more walkable, inclusive, and responsive urban environments. 
\end{abstract}

\section{Introduction}

Sidewalk delivery robots are a promising solution for deliveries in dense urban areas due to their potential to address congestion, sustainability, and operational efficiency \citep{jennings2019study, lemardele2021potentialities}. Robots can be dispatched from retail locations, mobility hubs, and even retrofitted vehicles for short-range trips and partially replace standard, less sustainable delivery methods. A relatively unexplored, yet potential benefit of sidewalk robots is their capacity to leverage navigation data (e.g., duration, speed, stops) and other trip-related measurements (e.g., pedestrian flows and behavior, pavement conditions) to investigate walkability-related features in cities.

Whether operational or perceived, walkability is increasingly recognized as a vital aspect of sustainable urban development. To develop more walkable cities, one must first unravel the complex pedestrian mobility patterns and interactions with existing infrastructure \citep{RHOADS2023101936}. Thus, increasing efforts are being made in assessing the usage of pedestrian infrastructure, identifying critical features such as pedestrian density (here referring to the number of pedestrians per unit area of sidewalk space) and pedestrian speed, to better understand the complex pedestrian mobility patterns and interactions \citep{fruin1970designing,maghelal2011walkability,dragovic2023literature}. Currently, pedestrian flow data is typically collected through infrared (IR) break-beam sensors and cameras placed at specific locations across the city \citep{moussaid2010walking, alahi2014robust,carter2020enhancing}. While this type of solution can offer granular data, it is typically difficult to adopt it on a broad scale due to its high costs and privacy issues (e.g., General Data Protection Regulation in Europe). More recently, street-level imagery (e.g., Google Street View) has been explored as an alternative source, where pedestrians can be detected and counted with computer vision techniques \citep{LIU2023102027,CHEN2020101481}. These imagery-based approaches are less costly to deploy and scalable across large areas, but they suffer from limited temporal resolution, since imagery is updated infrequently and cannot provide continuous measurements. Other studies focusing on the larger-scale analysis of urban mobility patterns (e.g., origins and destinations, mode choice, trip features) rely on GPS, wireless and cellular networks, and social media geotag records \citep{leslie2007walkability,kung2014exploring, Gabriele2020, gao2014detecting}. These solutions offer a good overview of pedestrian mobility at an aggregate level, but they cannot capture more complex phenomena such as, pedestrian flows, trajectories, and interactions with other road users. Surveys and audits are also used for collecting qualitative data on perceived walkability \citep{Abley2011, CAMBRA2020100797, CHEN2024102151} and quantitative measures of the built environment to develop walkability indicators \citep{BARTZOKASTSIOMPRAS2021107048, lee2020school}. Although these methods provide useful insights, they are time-consuming, resource-intensive, and difficult to scale, posing a major bottleneck to advancing the knowledge of walkability in research.  

Given these challenges, recent studies have explored the use of AI tools to enhance walkability assessment by improving cost efficiency and analytical depth. For instance, \citet{KI2025102319} proposed a framework that employs ChatGPT as an image processing agent to interpret multisource spatial data for walkability evaluation. Their approach demonstrates that ChatGPT can identify additional environmental features beyond the capabilities of conventional computer vision models. In a different direction, \citet{LIAN2024102057} utilized deep learning techniques to estimate pedestrian volumes from bus dashcam videos, providing a scalable method to support pedestrian-focused urban planning. These examples illustrate the growing potential of AI-driven approaches to overcome traditional limitations in walkability research. 

Building on this trend of leveraging emerging technologies for walkability analysis, sidewalk delivery robots, which are currently among the most common real-world applications of autonomous ground vehicles, offer a novel and scalable opportunity to address key data limitations by enabling continuous, real-time collection of sidewalk-level information. Recent studies have explored these robots from multiple perspectives, including their interactions and potential conflicts with pedestrians \citep{GEHRKE2023100789}, the regulatory challenges associated with mitigating disruptions and risks \citep{Kovacic08082024}, and their role in meeting the growing demand for local deliveries and e-commerce \citep{YANG2025104978}. Equipped with a suite of onboard sensors and real-time processing capabilities, these robots can autonomously collect and analyze data during routine deliveries. In this study, we illustrate how raw sensor data can be transformed into sidewalk walkability-related features at high spatial and temporal resolution. These features are categorized into three primary domains: robot trip features (e.g., robot speed, travel duration), sidewalk condition features (e.g., effective walkway width, unevenness), and sidewalk utilization features (e.g., pedestrian density and speed). Walkability is inherently multifaceted, shaped by social, cultural, and individual preferences. Although sidewalk robots cannot capture all factors influencing walkability, they can provide valuable insights into the walking infrastructure quality, spatio-temporal pedestrian patterns, and measurable aspects of the walking experience. 

This study does not aim to assess all variables related to walkability. Instead, it demonstrates how sidewalk robots can be leveraged to automatically capture measurable walkability-related features at the sidewalk level and use them to assess pedestrian mobility in practice. The contributions of this study are threefold. First, it introduces robots as a novel data collection instrument, capable of overcoming limitations of traditional methods  which prevent their application and scalability. As sidewalk delivery robots become more common in urban environments, their routine movements offer a practical and cost-efficient means to evaluate pedestrian mobility real-time. Second, by grouping different operational, walkability-related factors into robot trip features, sidewalk condition features, and sidewalk utilization features, the study establishes a systematic framework for extracting and analyzing data. Third, it provides empirical evidence that robot-based measurements can both confirm established relationships in pedestrian dynamics and reveal context-specific patterns, such as how width, density, and surface quality shape walking speed and movement styles. In the long run, this robot-based method can help build richer and more inclusive datasets that complement traditional audits and fixed sensors.

The remainder of this paper is structured as follows. In Section \ref{Sec:Literature Review}, we briefly review the relevant literature on features that influence sidewalk walkability. Section \ref{Sec:Data Collection} describes the data collection procedures and the methodology used to derive walkability-related features from raw robot sensor data. Section \ref{Sec:Analysis} includes several analyses of the extracted features, exploring differences in sidewalk conditions and their utilization, pedestrian behavior patterns, and the influence of sidewalk characteristics on pedestrian movement. Section \ref{Sec:discussion} offers deep reflections on our findings and discusses potential directions for future research. Finally, Section \ref{Sec:Conclusion} summarizes the main findings, highlights contributions. and outlines limitations and future research directions.

\section{Related Work}
\label{Sec:Literature Review}

In recent years, walkability has been extensively studied across diverse disciplines, resulting in a growing body of literature that examines both objective and perceived aspects of the walking environment. Several comprehensive reviews have synthesized these efforts. \citet{wang2019neighbourhood} categorized neighborhood walkability research into three major perspectives: walkability measurements, the built environment and health, and walkability application. \citet{fonseca2022built} focused on built environment attributes, identifying 32 key attributes evaluated through 63 specific measures. \citet{shields2023walkability} further expanded the scope by examining walkability across various contexts and highlighting the role of GIS-based tools in evaluating walkable environments. Across these studies, there is a consensus that built environment features significantly influence walkability. These features cover different aspects of the urban built environment, such as land use, accessibility, network connectivity, pedestrian facility and comfort, safety and security, and streetscape design \citep{fonseca2022built}. In addition to objective environmental factors, perceived walkability has also gained attention. For example, \citet{de2023determinants} created a conceptual model of perceived walkability, incorporating elements such as walking time, service accessibility satisfaction, infrastructure quality, perceived crime, and overall perception of walkability. 

While these studies offer a broad conceptual and empirical understanding of walkability, this study focuses on the operational features, specific environmental factors that directly affect the functional experience of walking. Rather than a systematic survey of the entire walkability literature, we summarize relevant across complementary fields, including urban planning, transport geography, pedestrian mobility studies, and traffic flow theory. The selected studies explicitly examined the relation between the built environment and pedestrian movements, provided measurable indicators at the sidewalk level, and offered empirical insights. On this basis , we identified twelve relevant factors from an operational perspective: 1) static obstacles; 2) dynamic obstacles; 3) walkway width; 4) walkway quality; 5) sidewalk slope; 6) crosswalk distance; 7) crosswalk type; 8) potential for conflicts; 9) vehicle speed; 10) traffic volume; 11) driveways; 12) supporting facilities. Among these, factors 6 to 12 are mainly related to crossing conditions. Each feature may correspond to multiple indicators or measurements across different studies. 

Obstacles in the walkway can cause deviation from the original path, which is proven to be one of the major factors that affect the walkability of a path according to \citet{Abley2011}.  \citet{Reisi2019LocalWI} applied the Analytical Hierarchy Process (AHP) for weighting indicators of walkability. The number of static obstacles (e.g. poles, signs, chairs along sidewalks) was found to have high relative weight. Other static obstacles, such as wayfinding, green hedge, trees, trash cans, public seating furniture, and illegally parked vehicles, were also considered to have an influence on walkability \citep{YIN2017288,MITROPOULOS2023103566,BARTZOKASTSIOMPRAS2021107048}.  In addition to using the number of obstacles as a measure, \citet{Golan_Wilkinson_Henderson_Weverka_2019} employed the number of requests for cleanup of streets, garbage cans, bulky items, and other waste, which was ranked as the third most influential variable on women's walkability. Beyond the obstacles' general impact on route continuity and convenience, they can pose significant hazards for people with disabilities, particularly blind or visually impaired pedestrians. 

Aside from static obstacles, the impact of dynamic obstacles on walkability should not be underestimated. Dynamic obstacles refer to temporary or moving obstructions that interrupt or constrain the walking experience. These may include illegally parked cars, delivery vehicles making curb stops, scattered e-scooters and bicycles, street vendors or construction equipment occupying sidewalks, and, most commonly, other pedestrians. Several studies have examined how pedestrian presence affects walkability. \citet{fruin1970designing} and \citet{Gabriele2020} used pedestrian flow to describe the pedestrian level of service (LOS), which is an indicator of walkability. \citet{Arellana2020} incorporated pedestrian volume as one of the components in assessing the safety factor when measuring walkability. The observed variation in pedestrian volume can also be utilized to indicate or symbolize the alteration in walkability \citep{CAMBRA2020100797}. Moreover,  the results of \citet{JARDIM2023101993} revealed that the strength of the relationship between pedestrian flow and walkability differs between weekdays and weekends. 

Walkway width is highly appropriate as an indicator, as it directly influences the amount of space available for pedestrian movement. \citet{moura2017measuring} considered effective walkway width as one of the key concerns in the convenience factor in walkability scoring. \citet{Abley2011} found that the minimum effective width along the path has a positive relationship with walkability. Moreover, the average walkway width along the segment can also be used as a measure. Wider sidewalks are generally rated to be more walkable, and the difference between a narrow and a wide walkway has a greater navigational impact than minor differences between two already wide walkways on sidewalk robots \citep{CORNO202015053}. Despite its importance, effective width is often difficult to measure directly. A practical alternative is to use street audit questions, such as `Is the net width of the sidewalk supportive for three individuals to walk simultaneously and in parallel?', according to  \citet{BARTZOKASTSIOMPRAS2021107048}.

The walkability improves as the physical road conditions enhance \citep{Abley2011}, thus the quality of the walkway can be qualitatively evaluated based on regularity, smoothness, and slippery characteristics of the pavement \citep{CORNO202015053,moura2017measuring}, and is often evaluated through field audits \citep{BARTZOKASTSIOMPRAS2021107048,lee2020school}. In addition to these attributes, pavement friction and texture are also important characteristics influencing pedestrian movement. Together, these surface properties not only shape comfort and accessibility but also play a critical role in safety, as poor conditions such as uneven or slippery pavements can increase the risk of pedestrian falls \citep{silva2025engineering}. Closely related to surface quality is the sidewalk slope, which affects both walking comfort and accessibility, especially for individuals with mobility limitations. Slope can be measured in two primary ways: by calculating the average gradient along a segment \citep{Deng2020,taleai2018integration}, or by measuring the elevation range between the highest and lowest points on the segment \citep{Fan2018WalkabilityIU}. 

When it comes to crosswalks, their length and type have been considered in walkability-related studies. The crosswalk length feature was often measured directly by the length of crosswalks \citep{Abley2011}, while the number of traffic lanes can be used instead in audit investigations \citep{BARTZOKASTSIOMPRAS2021107048}. Crosswalk type can be defined in different ways according to different case studies, such as the presence of crosswalk \citep{Bracy2014,lee2020school} and pedestrian signals \citep{Bracy2014,moura2017measuring}. \citet{CORNO202015053} categorized crosswalk type into five distinct groups (1. crosswalks with dedicated pedestrian traffic light; 2. crosswalks with dedicated pedestrian traffic light at intersections; 3. crosswalks without traffic light; 4. crosswalks without traffic light at an intersection; 5. crosswalks without zebra crossings) based on the presence of dedicated infrastructure. 

Safety-related features are very important, especially in certain situations such as uncontrolled crossings and mixed lanes for pedestrians, bicycles, and cars. They can include the potential for conflicts, vehicle speed, traffic volume, and the presence of driveways. One way to measure the potential for conflicts is to calculate the number of potential conflict points with bicycles, motor bikes, and cars at crosswalk \citep{CAMBRA2020100797,moura2017measuring}. The presence of segregation between pedestrian and vehicle routes can also be used as a measurement because it reduces the number of conflicts between pedestrians and vehicles on the street \citep{harun2020walkability}.  While there is no segregation at the intersection of sidewalks and driveways, in this case the number of driveways crossing the sidewalk to enable access to buildings constitutes a relevant factor of risk on walkability \citep{CORNO202015053}.

When dealing with uncontrolled crossings, \citet{Abley2011} considered the average vehicle speed of the segment as one of the key factors that significantly affect walkability. \citet{Golan_Wilkinson_Henderson_Weverka_2019} found that the maximum vehicle speed limit was ranked as the fourth most influential variable on women's walkability. Additionally, the traffic volume, measured by the number of vehicles, is also related to the safety feature of walkability \citep{williams2018neighborhood}. However, studying the impact of traffic volume on walkability is usually done at the macro level, such as calculating the total traffic volume or the average traffic volume per unit distance of a certain area's road network, which can reflect the attractiveness and connectivity of the area to a certain extent \citep{Billie2011}.

Supporting facilities encompass elements such as ramps or steps at the intersection of sidewalks and crosswalks, central islands within crosswalks, and sidewalk buffers. Ramps facilitate accessibility and are particularly beneficial for individuals with disabilities \citep{mulyadi2022walkability}, whereas steps are often considered barriers, especially for older adults, wheelchair users, or parents with strollers. Additionally, the presence of central islands and buffers is crucial for ensuring walkway safety, which in turn contributes to overall walkability \citep{Bracy2014,BARTZOKASTSIOMPRAS2021107048,Abley2011}.

In addition to the twelve identified operational features, some external environment-related features, including weather, temperature, and light conditions, can also be considered. Various weather conditions, such as rain, wind, snow, and sunny days, can have varying effects on walkability \citep{Abley2011,harun2020walkability,ALSHAMMAS2023101981}. The land surface temperature may affect pedestrians’ willingness to travel and the performance of robots, such as overheating (high temperature) and battery capacity reduction  (low temperature). \citet{Savvides2016} reported that an increase in air temperature of 10°C results in a 15\%-20\% increase in the proportion of individuals opting to walk in the shade. It can be represented by Physiological Equivalent Temperature (PET) or calculated by satellite's thermal bands \citep{taleai2018integration}. Similarly, the availability of shade and protection from solar heat can be considered as features, as they have been demonstrated to be highly significant factors contributing to the walkability of streets \citep{harun2020walkability}. The light condition feature can be accessed by either counting the number of lights \citep{Reisi2019LocalWI,lee2020school} or by measuring the lumen. As reported by \citet{harun2020walkability}, it has a positive correlation with travel safety.

\begin{figure}[t]
    \centering
        \includegraphics[width=0.7\linewidth]{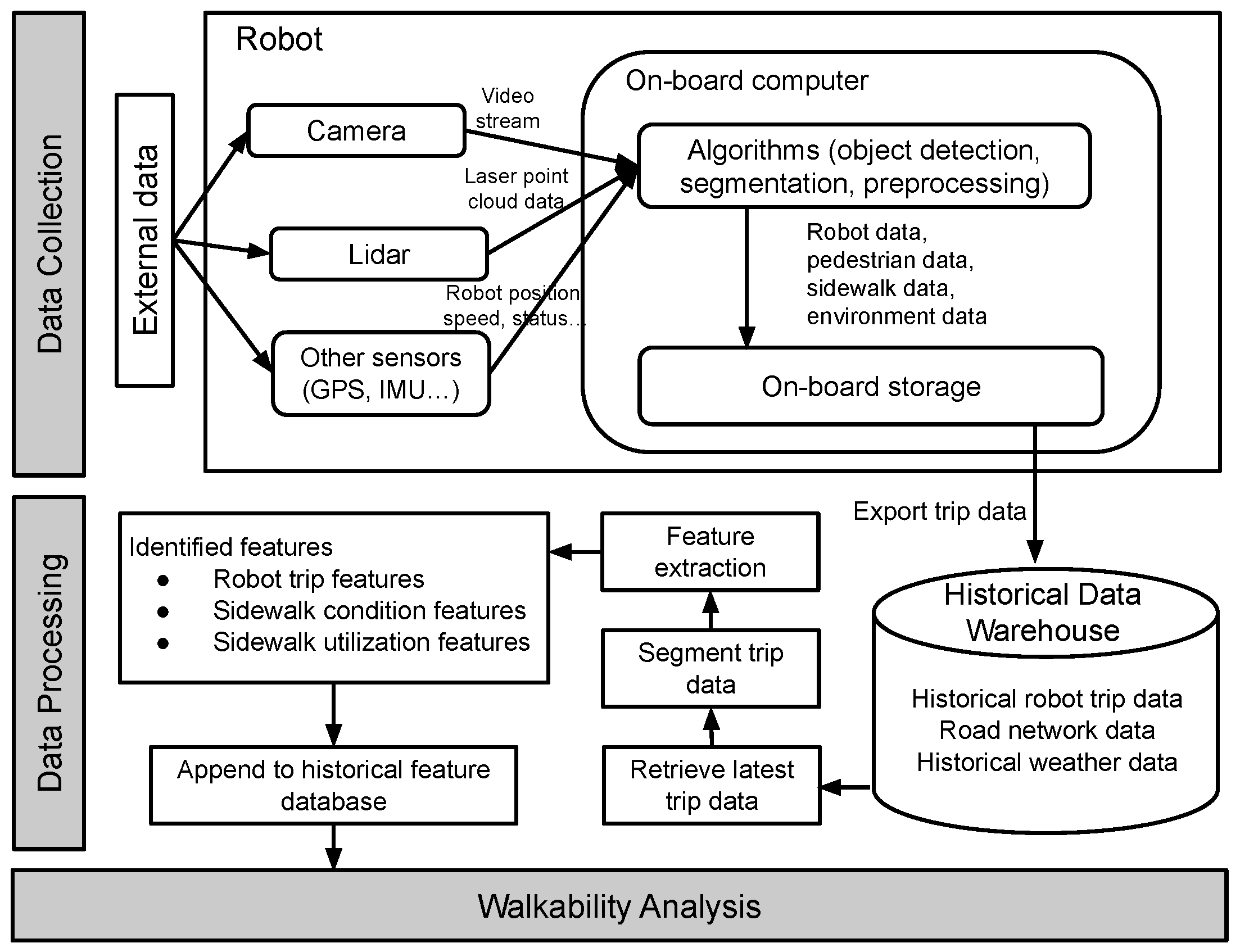}
    \caption{Work flow of sidewalk robot-based data collection and processing for walkability analysis}
    \label{fig:1}
\end{figure}

\section{Walkability-Related Features}
\label{Sec:Data Collection}

This section outlines the data collection process and the methodology used to derive walkability-related features from raw robot sensor data. We first describe the experimental setup, including the sidewalk robot platform and its deployment across the sidewalk network. The subsequent subsections present three key categories of features extracted from the collected data: robot trip features that characterize movement behavior, sidewalk condition features that reflect surface quality and geometry, and sidewalk utilization features that capture patterns of pedestrian interaction with the environment. Figure~\ref{fig:1} provides an overview of the data collection workflow and processing pipeline, illustrating how these components contribute to the walkability analysis.

\subsection{Experimental setup}

\begin{figure}[t]
    \centering
        \includegraphics[width=0.7\linewidth]{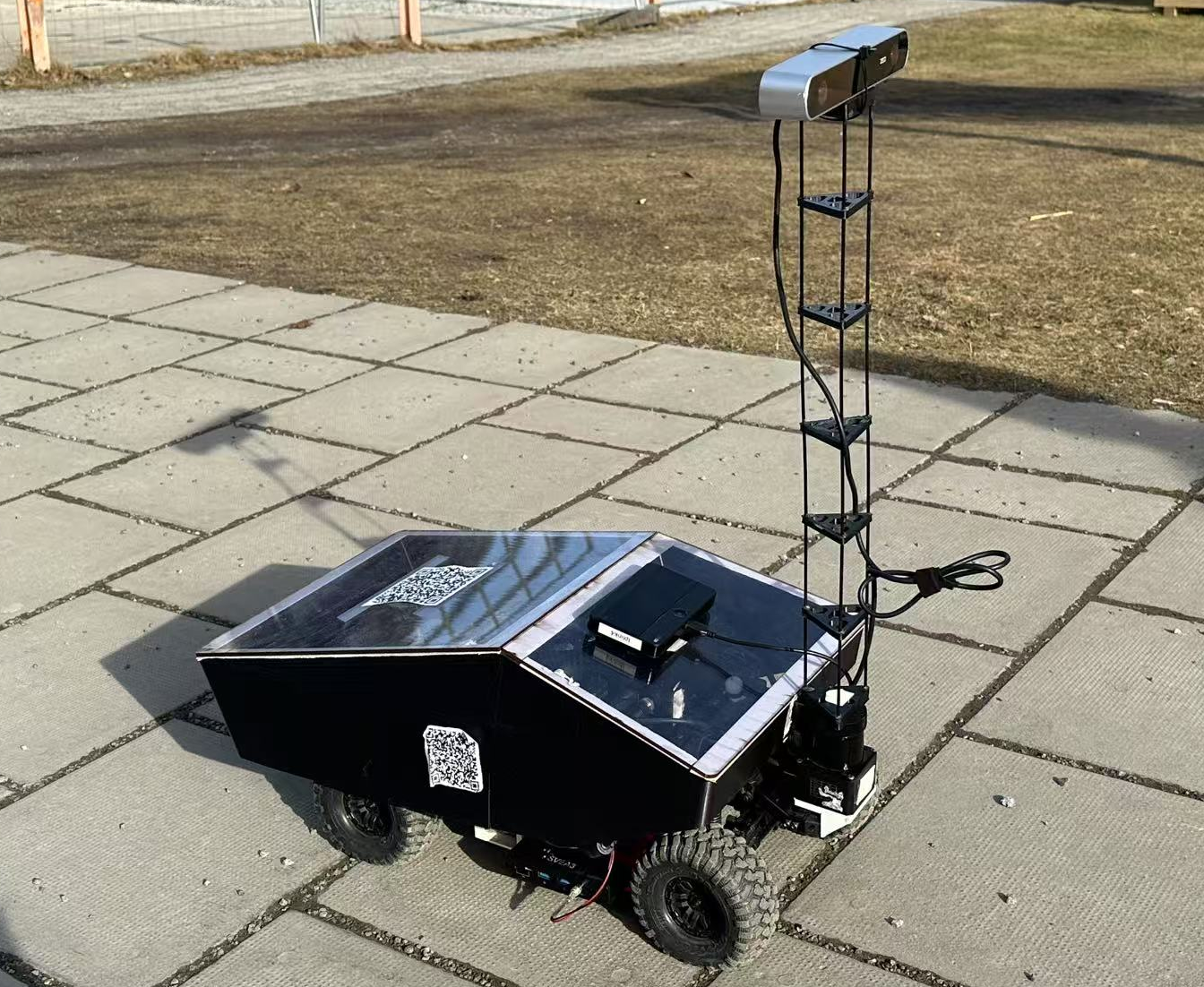}
    \caption{SVEA robot}
    \label{fig:2}
\end{figure}

The robot utilized for sidewalk data collection is a custom-designed, open-source vehicle called "Small Vehicles for Autonomy" (SVEA) \citep{svea}, as shown in Figure~\ref{fig:2}. Although the testbed of SVEA is designed to support automated driving systems under V2X communication, in this study, the robot was manually driven along sidewalks to collect data for safety reasons. The platform is built on the Traxxas TRX-4, a four-wheel-drive vehicle equipped with a two-gear transmission system. To operate the robot in a manner similar to a typical sidewalk autonomous delivery robot, which travels at pedestrian speeds, we selected the low-gear mode and programmatically restricted its velocity during the data collection experiment. With these settings, the maximum speed of the robot in low gear was capped at approximately 1.6 m/s. The robot's footprint measures 28 cm × 57 cm, and its weight is 6.5 kg. It is equipped with a suite of onboard sensors, including a Stereolabs ZED 1 stereo camera, a u-blox ZED-F9P GNSS module for RTK-GPS, a Hokuyo UST-10LX LiDAR, and an off-the-shelf Adafruit IMU, to achieve real-time perception and localization. On-board computing is facilitated by the Stereolabs ZED-BOX, which contains a Jetson Xavier NX, serving as the central computing unit that processes sensor data in real time. 

The stereo camera is mounted on an extended frame to elevate and improve its field of view. This configuration brings the robot's total height to 68 cm where the camera is located. The video stream is fed from the camera into the computing unit, where its built-in algorithm processes it in real time into desensitized data for storage. The pedestrians are detected using Ultralytics' off-the-shelf YOLOv8 model \citep{ultralytics2023yolov8} in each camera frame, resulting in a bounding box. Their positions in the 3D space are also estimated together with the depth map produced by the camera. At the same time, the images are segmented using FastSAM, a CNN Segment Anything Model \citep{zhao2023fastsam}. The sidewalk segmentation is related to the point cloud generated by the camera. Through this, which points in the point cloud relate to the sidewalk can be determined, and consequently the sidewalk width. This setup allows us to perform global localization of the robot and perceive the surrounding environment, enabling data collection of pedestrian motions in relation to the robot and in dynamic sidewalk environments.

Furthermore, in compliance with GDPR requirements, the robot does not store any raw image or video data during operation. All visual illustrations included in this paper were captured independently for demonstration purposes and are carefully curated to exclude any personally identifiable information.

\begin{figure}[t]
    \centering
    \begin{subfigure}{0.47\textwidth}
        \centering
        \includegraphics[width=\linewidth]{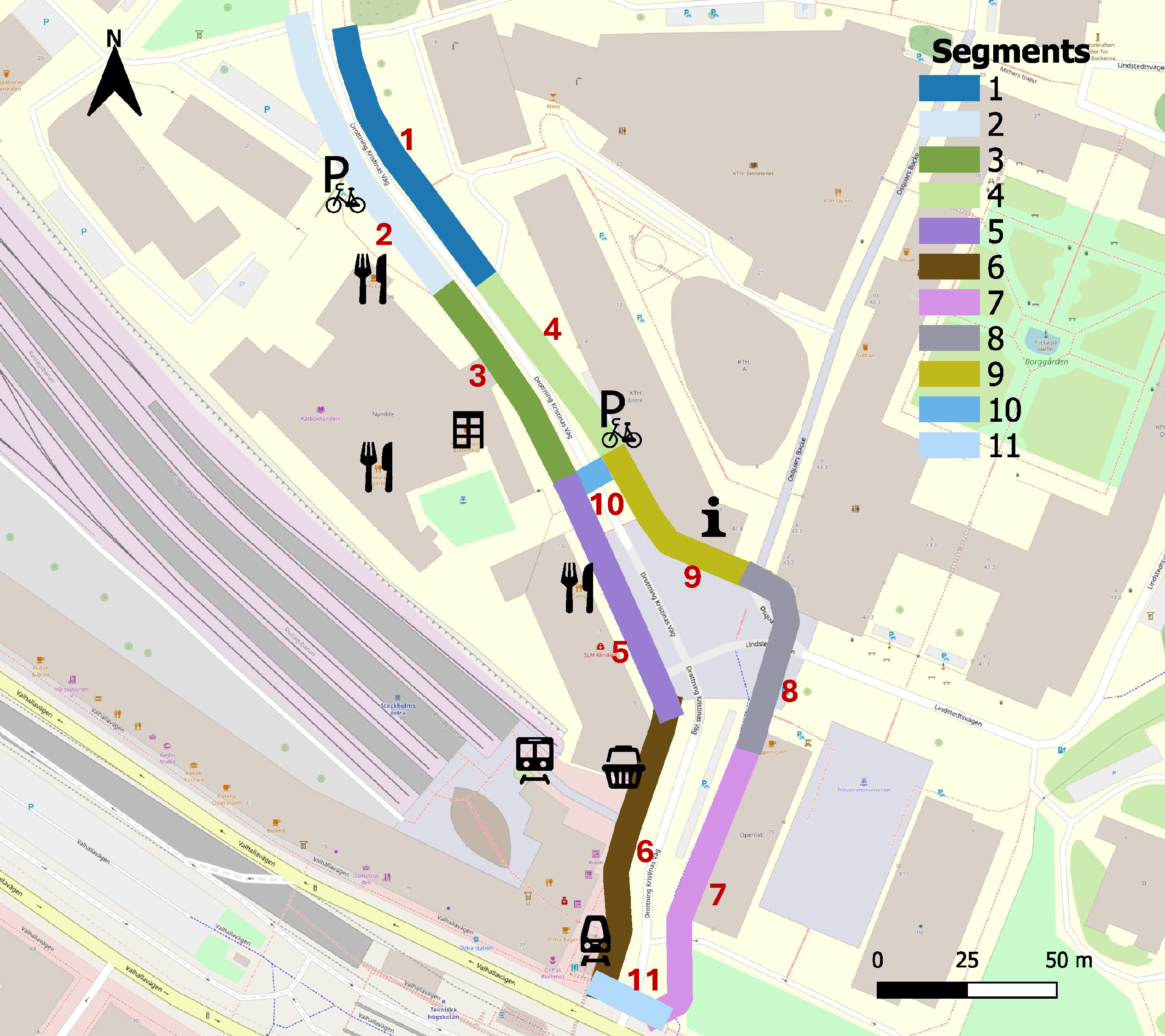}
        \caption{Sidewalk network}
        \label{fig:3a}
    \end{subfigure}
    \hfill
    \begin{subfigure}{0.52\textwidth}
        \centering
        \includegraphics[width=\linewidth]{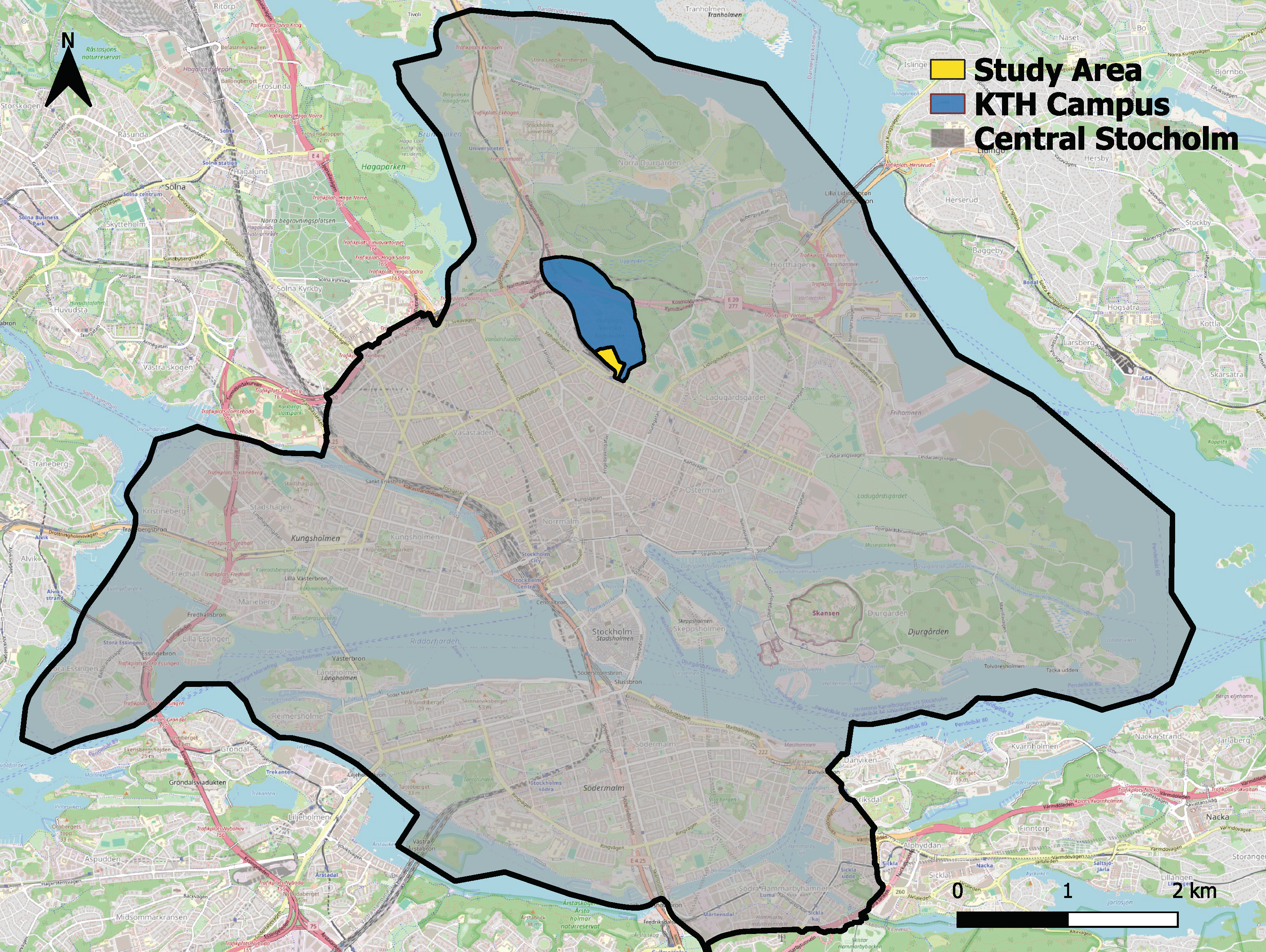}
        \caption{Network location within Stockholm}
        \label{fig:3b}
    \end{subfigure}
    \caption{Network used in the experiments on KTH campus}
    \label{fig:3}
\end{figure}

The robot was deployed on a small sidewalk network within the KTH campus in Stockholm, Sweden, as shown in Figure~\ref{fig:3}. The network consists of multiple segments, including both sidewalks and pedestrian crossings, designed to capture a variety of walking environments. For each street in the selected area, both sidewalks were included in the analysis rather than limiting the study to a single side. As illustrated in Figure~\ref{fig:3a}, the sidewalks on the southwest side have more points of interest compared to those on the northeast side, resulting in different levels of pedestrian activity across the network. Each segment therefore serves as a unit of analysis for evaluating sidewalk features across the area.

KTH campus was selected as the study area due to several reasons. As an urban campus, it offers a diverse range of pedestrian facilities, and stable pedestrian volumes. The selected area, which exhibits high network centrality within the campus sidewalk system, is traversed daily by most students and staff, making it an high-activity pathway. Its combination of academic buildings, green spaces, and service facilities generates periodic pedestrian flows. These characteristics create dynamic settings that support realistic testing, while also offering a manageable environment for repeated robot deployments. Safety and practicality also played a role since the robot needed to be supervised during the experiments. A concise overview of the network characteristics is provided in Table~\ref{tab:1} to support interpretation and transferability. 

\begin{table}[ht]
\centering
\caption{Summary of sidewalk network characteristics in the KTH campus study area.}
\label{tab:1}
\begin{tabular}{p{3.5cm} p{9.5cm}}
\hline
\textbf{Category} & \textbf{Description} \\
\hline
Scale and composition & Small-scale network including both sidewalks for each street, and two pedestrian crossings.\\
Activity and usage & High centrality within campus; daily flows of students and staff; periodic peaks during class hours; southwest side with higher activity due to points of interest. \\
\end{tabular}
\end{table}

From September 18, 2024, to March 24, 2025, the robot completed 101 trips, covering 900 records of sidewalk segments in total. Trips were conducted on different days and at various times between 08:00 and 18:00, with particular focus on peak hours. During each trip, onboard sensors continuously recorded raw data, including GNSS points, robot velocity and heading, IMU readings, sidewalk width, ambient brightness, and anonymized pedestrian data. The experiments were conducting during months of higher pedestrian activity on campus, ensuring the dataset’s representativeness. 

The recorded raw data from each trip were stored in a centralized data warehouse that maintains all historical trip records. Each trip was divided into sidewalk segments, and segment-level features were extracted and grouped into three categories: robot trip features, sidewalk condition features, and sidewalk utilization features. These extracted features offer valuable insights into the quality of walking infrastructure and spatio-temporal pedestrian patterns. An overview of these features is presented in Table~\ref{tab:2}, while an interactive dashboard\footnote{\url{https://ismir-sidewalk-mobility-data-dashboard.streamlit.app/}} visualizes sidewalk usage and characteristics for the study area. In the subsequent analyses (Section \ref{Sec:Analysis}), these features are used in two ways: (i) segment-wise averages are computed to compare conditions and utilization across the network; and (ii) the individual segment-level records are analyzed directly for box-plot visualization, Pearson correlation, clustering, and regression.

\begin{table}[ht]
        \centering
        \caption{Overview of the identified features}
        \footnotesize
        \renewcommand{\arraystretch}{1.2} 
        \begin{tabular}{|l|l|c|} \hline 
        
        \textbf{Feature Name}&
        \textbf{Description}&
        \textbf{Math notation}\\ \hline 

        \multicolumn{3}{|c|}{\textbf{Robot Trip Features}} \\ \hline  
        \texttt{segment\_duration} (s)& Time taken to traverse this segment.&\(T_{\text{seg}}\)\\ \hline 
        \texttt{segment\_distance} (m)& Distance traveled in this segment.&\(D_{\text{seg}}\)\\ \hline 
        \texttt{relative\_duration} (s)& Ratio of segment duration to its shortest recorded duration.&\(T_{\text{seg}}^{\text{r}}\)\\ \hline 
        \texttt{relative\_distance} (m)& Ratio of segment distance to its shortest recorded distance.&\(D_{\text{seg}}^{\text{r}}\)\\ \hline 
        \texttt{segment\_max\_speed} (m/s)& Maximum robot speed in this segment.&\(v_{\max}^{\text{seg}}\)\\ \hline 
        \texttt{segment\_min\_speed} (m/s)& Minimum robot speed in this segment.&\(v_{\min}^{\text{seg}}\)\\ \hline 
        \texttt{segment\_avg\_speed} (m/s)& Average robot speed in this segment.&\(v_{\text{avg}}^{\text{seg}}\)\\ \hline 
        \texttt{speed\_drop\_avg} (m/s²)& Average deviation from desired speed.& \(v_{\text{drop}}\)\\ \hline 
        \texttt{trip\_peak\_speed} (m/s)& Peak speed of the robot during the trip.&\(v_{\text{peak}}\)\\ \hline 
        \texttt{num\_stops}& Number of times the robot stopped in this segment.&\(n_{\text{stops}}\)\\ \hline 
        \texttt{total\_wait\_time} (s)& Total waiting time in this segment.&\(T_{\text{wait}}\)\\ \hline 

        \multicolumn{3}{|c|}{\textbf{Sidewalk Condition Features}} \\ \hline  
        \texttt{segment\_length} (m)& Length of the segment.&\(L_{\text{seg}}\)\\ \hline 
        \texttt{segment\_width} (m)& Width of the segment.&\(W_{\text{seg}}\)\\ \hline 
        \texttt{min\_effective\_width} (m)& Minimum effective width available.&\(W_{\min}^{\text{eff}}\)\\ \hline 
        \texttt{avg\_effective\_width} (m)& Average effective width available.&\(W_{\text{avg}}^{\text{eff}}\)\\ \hline 
        \texttt{segment\_slope}& Measured slope of the sidewalk segment.&\(\theta_{\text{seg}}\)\\ \hline 
        \texttt{lighting\_condition}& Brightness level in this segment.&\(L_{\text{cond}}\)\\ \hline 
        \texttt{irregularity\_index}& Index measuring sidewalk irregularities.&\( I_{\text{seg}} \)\\ \hline  
        \texttt{unevenness\_index}& Index measuring sidewalk unevenness.&\( U_{\text{seg}} \)\\ \hline 

        \multicolumn{3}{|c|}{\textbf{Sidewalk Utilization Features}} \\ \hline  
        \texttt{total\_ped\_count}& Total number of pedestrians detected.&\(N_{\text{ped}}\)\\ \hline 
        \texttt{avg\_ped\_speed} (m/s)& Average pedestrian speed.&\(v_{\text{avg}}^{\text{ped}}\)\\ \hline  
 \texttt{ped\_speed\_variation} (m/s)& Standard deviation of pedestrian speed.&\(\sigma_{\text{avg}}^{\text{ped}}\)\\ \hline  
 \texttt{ped\_turns}& Number of turning angles exceeding 30 degrees.&\(n_{\text{turns}}\)\\ \hline  
 \texttt{ped\_path\_deviation} (m)& Average distance pedestrians deviate from a straight path.&\(\delta_{\text{path}}\)\\ \hline 
        \texttt{max\_ped\_density} (Peds/m²)& Maximum pedestrian density.&\(k_{max}\)\\ \hline 
        \texttt{avg\_ped\_density} (Peds/m²)& Average pedestrian density.&\(k_{avg}\)\\ \hline
        \end{tabular}
    \label{tab:2}
    \end{table}

\subsection{Robot trip features}

The robot trip features describe the movement characteristics of the robot while traversing each sidewalk segment. These features provide insights into how the robot interacts with the pedestrian environment during its trips.

Since segments vary in length and traversal time, direct comparisons between segments can be misleading. To address this, segment duration (\(T_{\text{seg}}\)) and segment distance (\(D_{\text{seg}}\)) are normalized using the shortest recorded values for each segment across all trips. This results in the features, relative duration and relative distance, which indicate how much longer or farther a segment traversal is relative to its minimum values.

Although the maximum speed of the robot in low gear was constrained to approximately 1.6 m/s, variations in battery performance caused fluctuations in the actual maximum speed across trips. Thus, the highest speed recorded in the trip, denoted as \(v_{\text{peak}}\), is calculated to quantify this variability. This helps identify potential data inconsistencies due to variations in battery output, which can affect movement performance. 

To further assess the influence of sidewalk conditions on the robot's movement, while excluding variations caused by battery inconsistencies, we measured the average deviation from the desired speed, \(v_{\text{drop}}\), defined as:
\begin{equation}\label{eq:3.2.1}
v_{\text{drop}} = \frac{1}{T_{\text{seg}}} \int_0^{T_{\text{seg}}} \big( v_{\text{peak}}- v(t) \big) dt
\end{equation}
where \(v(t)\) is the instantaneous speed of the robot at time \(t\) during the segment. A higher \(v_{\text{drop}}\) value indicates more substantial slowdowns, potentially caused by surface irregularities, obstructions, or pedestrian congestion that may reduce effective walkability.

\subsection{Sidewalk condition features}

\begin{figure}
    \centering
    \begin{subfigure}{0.5\textwidth}
        \centering
        \includegraphics[width=\linewidth]{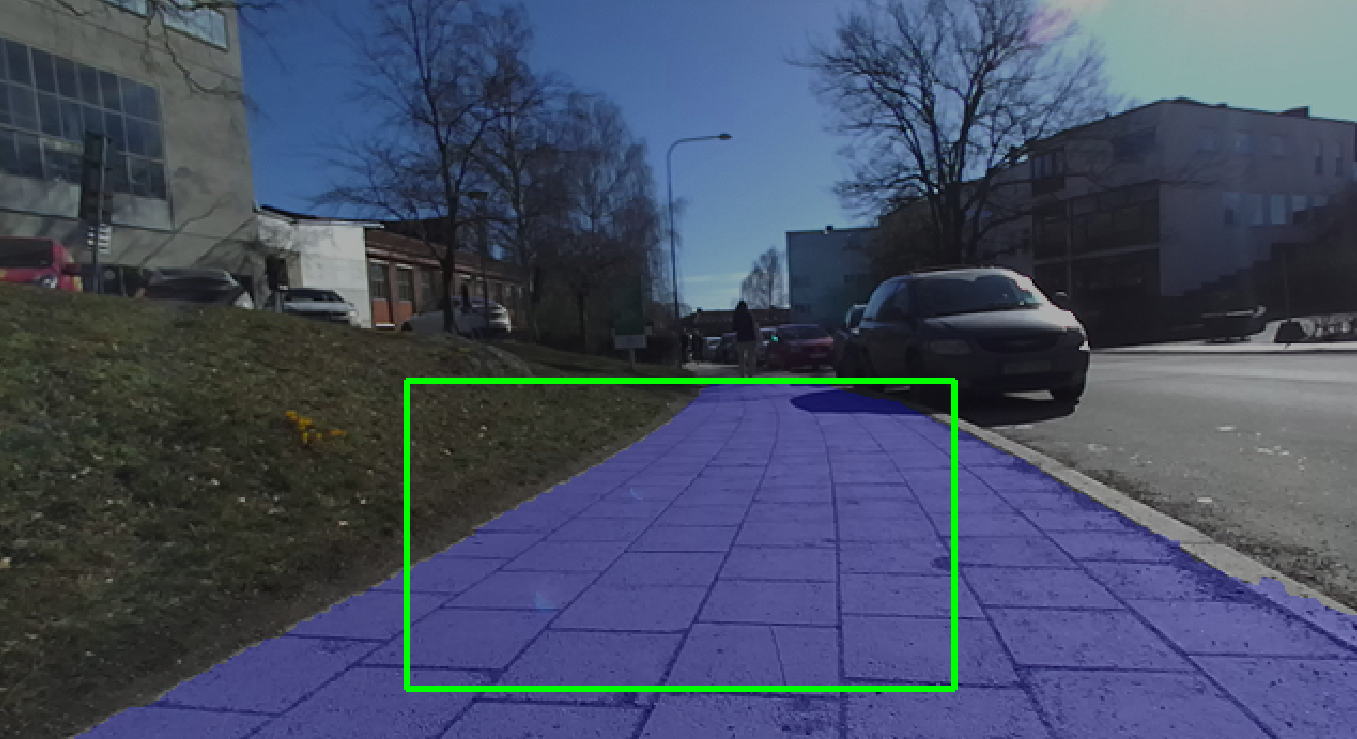}
        \caption{Segmentation}
        \label{fig:4a}
    \end{subfigure}
    \hfill
    \begin{subfigure}{0.49\textwidth}
        \centering
        \includegraphics[width=\linewidth]{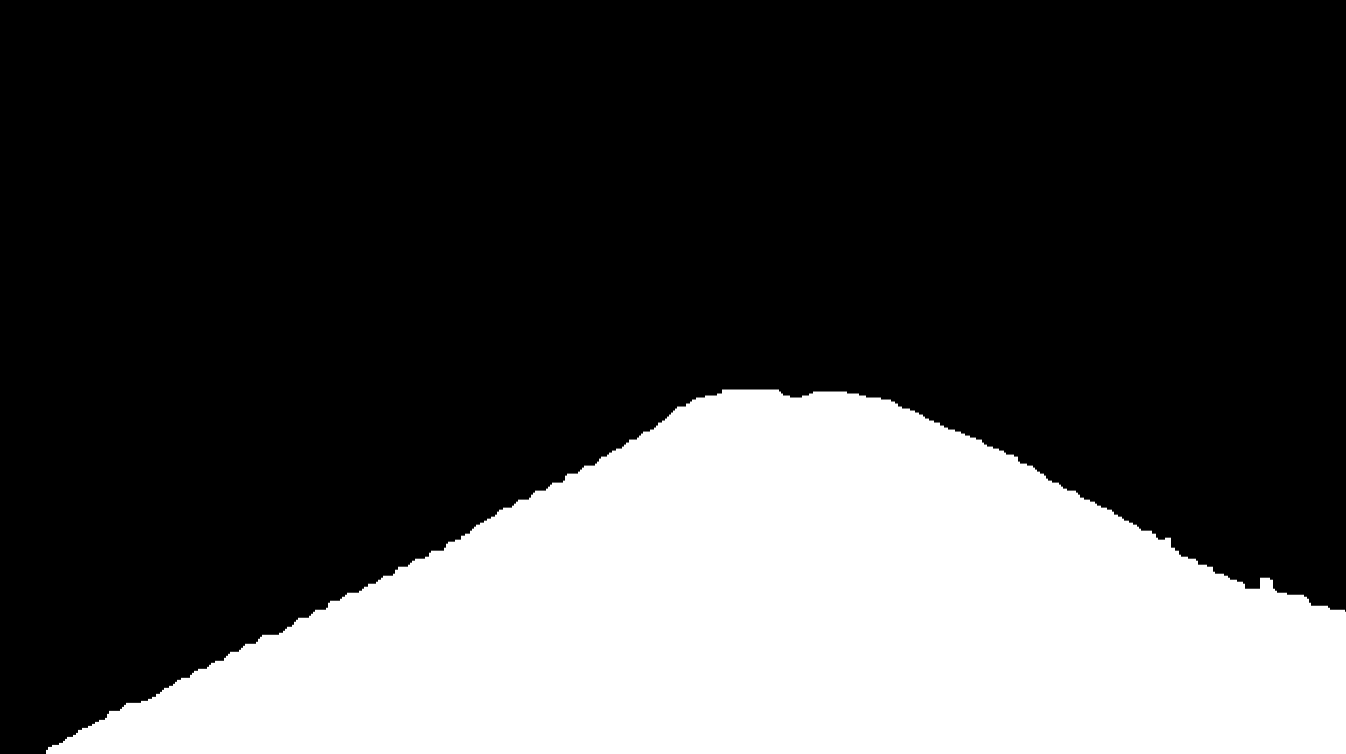}
        \caption{Segmentation masked}
        \label{fig:4b}
    \end{subfigure}
    \caption{Sidewalk segmentation}
    \label{fig:4}
\end{figure}

The sidewalk condition features describe the physical state and accessibility of sidewalks and crossings, including segment distance, width, slope, brightness, and quality of the pavement. The effective sidewalk width at a 2-meter distance ahead of the robot is computed using the point cloud data related to the sidewalk segmentation (Figure~\ref{fig:4}). To determine the effective width, the 3D coordinates of the sidewalk boundaries at the 2-meter mark are first extracted. The effective width is then derived as the Euclidean distance between the left and right sidewalk boundary points in the robot's coordinate frame. This measurement accounts for occlusions, obstacles, and environmental constraints that are commonly encountered during real-life sidewalk trips, which can provide an accurate representation of the walkable sidewalk space. This data is collected at a frequency of one measurement per second. Based on the effective width along the entire segment, the average effective width is then derived.

The slope feature for each sidewalk segment is derived from the altitude data collected by the robot during its traversal. The robot records the elevation at both the origin and destination points of each segment. Slope is then calculated as the difference in altitude between the destination and origin, divided by the segment length. A positive slope value indicates an uphill segment, while a negative value represents a downhill segment. This metric captures the gradient of the sidewalk surface, which can influence both pedestrian movement and robot mobility.  

Sidewalk quality is assessed by evaluating the irregularities and unevenness of the sidewalk surface in this study, as these factors can significantly impact pedestrian comfort and sidewalk accessibility. Other important characteristics, such as pavement friction and texture, were not collected and are left for future research. Irregularities refer to sudden and localized disruptions on the sidewalk, such as cracks, bumps, or abrupt height changes, which can cause discomfort or pose challenges for both pedestrians and autonomous robots. Unevenness, on the other hand, represents more gradual variations in the sidewalk surface, such as inclines, depressions, or wavy patterns, which may influence walking stability over a longer distance. In this study, both irregularities and unevenness were derived from vertical acceleration data obtained from the IMU sensor. While this unified sensing approach ensures consistency across features, it should be noted that irregularities and unevenness are conceptually distinct and are often measured using different methodologies. Relying solely on vertical acceleration may therefore introduce limitations in accuracy, and this should be considered when interpreting the results.

Irregularities are detected through an event-based approach, and processed into the Sidewalk Surface Condition Index, as proposed by \citet{CORNO202015053}. The detection process consists of a signal processing pipeline designed to isolate and quantify irregularities in the sidewalk surface. First, a high-pass filter is applied to the vertical acceleration in order to eliminate gravity and low-frequency components associated with the robot's pitch dynamics. Next, an overlapping time sliding window of 1 second with a 0.1-second step is used to compute the root mean square (RMS) of the filtered acceleration \(a_z^{\text{high}}(t)\) within each window. For a window of duration \( T_w \), the RMS is computed as:
\begin{equation}\label{eq:3.3.1}
a_{\text{RMS}}^{\text{high}}(t) = \sqrt{\frac{1}{T_w} \int_{t}^{t+T_w} \left( a_z^{\text{high}}(\tau) \right)^2 d\tau}
\end{equation}

An irregularity event is detected when \( a_{\text{RMS}}^{\text{high}}(t)\) exceeds a predefined threshold. Since vertical acceleration is influenced by the robot's speed, the irregularity value for each event is normalized by the robot's velocity \(v(t)\) at the time of detection. The resulting normalized irregularity value, denoted as \( I(t) \), is defined as:
\begin{equation}\label{eq:3.3.2}
I(t) = \frac{a_{\text{RMS}}^{\text{high}}(t)}{v(t)}
\end{equation}

However, irregularities are not uniformly distributed along the sidewalk, meaning that the robot may detect them during some trips but miss them during others. To derive a segment-level irregularity measure, detected events from multiple trips are overlapped and clustered based on their spatial proximity. The normalized irregularity values within each cluster are averaged, resulting in a representative value \( I_i \) for each cluster. The spatial extent of each cluster is then mapped back to the corresponding sidewalk segments. The final segment-level \texttt{irregularity\_index}, denoted as \( I_{\text{seg}} \), is computed by summing the contributions of all clusters within the segment, weighted by their lengths:
\begin{equation}\label{eq:3.3.3}
I_{\text{seg}} = \sum_{i \in \mathcal{C}} I_i \cdot L_i
\end{equation}
where \(\mathcal{C}\) is the set of irregularity clusters within the segment, and \(L_i\) is the length of the cluster mapped on the segment. This feature is computed once per segment and remains consistent across trips. The resulting irregularity index is a non-negative continuous value, where values closer to 0 indicate smoother, less disrupted sidewalks, and larger values indicate more severe or more spatially extensive irregularities.

Unevenness measurement, unlike irregularities, is measured continuously rather than in discrete events. It captures smoother, longer-wavelength variations in the sidewalk surface. To compute this feature, the vertical acceleration is first passed through a low-pass filter, instead of a high-pass filter, to suppress high-frequency noise from motor vibrations, small bumps, and sensor errors. The filtered signal, \(a_z^{\text{low}}(t)\), is then normalized by the robot's velocity \(v(t)\) at the time of measurement, to account for velocity-dependent variations:
\begin{equation}\label{eq:3.3.4}
\hat{a}_z(t) = \frac{a_z^{\text{low}}(t)}{v(t)}
\end{equation}

The RMS of the normalized signal is computed over the entire segment to quantify unevenness:
\begin{equation}\label{eq:3.3.5}
a_{\text{RMS}}^{\text{norm}} = \sqrt{\frac{1}{T_{\text{seg}}} \int_0^{T_{\text{seg}}} \big( \hat{a}_z(t) \big)^2 dt}
\end{equation}

Finally, the unevenness, denoted as \( U_{\text{seg}} \), is computed as the average of \( a_{\text{RMS}}^{\text{norm}} \) across multiple trips recorded for the same segment:
\begin{equation}\label{eq:3.3.6}
U_{\text{seg}} = \frac{1}{N_{\text{trips}}} \sum_{j=1}^{N_{\text{trips}}} a_{\text{RMS},j}^{\text{norm}}
\end{equation}

Like the irregularity index, the unevenness measure is a non-negative continuous value, with values closer to 0 reflecting a smoother and more stable surface, and larger values representing sidewalks with greater long-wavelength unevenness. 

Although Eq.\eqref{eq:3.2.1} and Eq.\eqref{eq:3.3.5} are written in continuous form for clarity, in practice, all computations are performed over discretely sampled data (e.g., at 1 Hz) using numerical approximations. Future work could integrate complementary sensing methods, such as LiDAR or vision-based detection, to enhance the robustness and accuracy of irregularity and unevenness measurements.

\subsection{Sidewalk utilization features}

Sidewalk utilization features capture the presence and movement of pedestrians within each segment, providing insights into how the sidewalk space is used. These features include the pedestrian total count, average speed, speed variation, frequent turning, path deviation, and density metrics, such as maximum pedestrian density and average pedestrian density. They can help assess sidewalk congestion levels and offer valuable information for evaluating the walkability. 

\begin{figure}
    \centering
    \begin{subfigure}{0.47\textwidth}
        \centering
        \includegraphics[width=\linewidth]{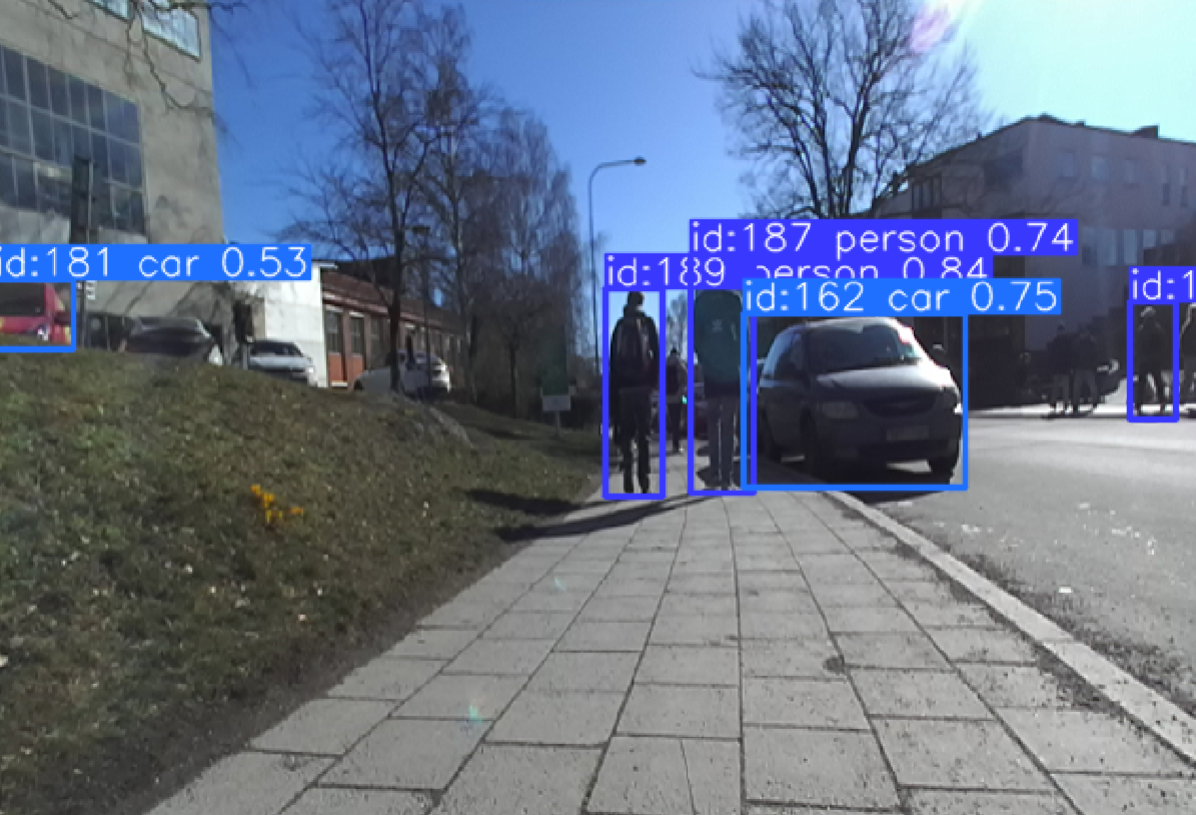}
        \caption{Object detection}
        \label{fig:5a}
    \end{subfigure}
    \hfill
    \begin{subfigure}{0.51\textwidth}
        \centering
        \includegraphics[width=\linewidth]{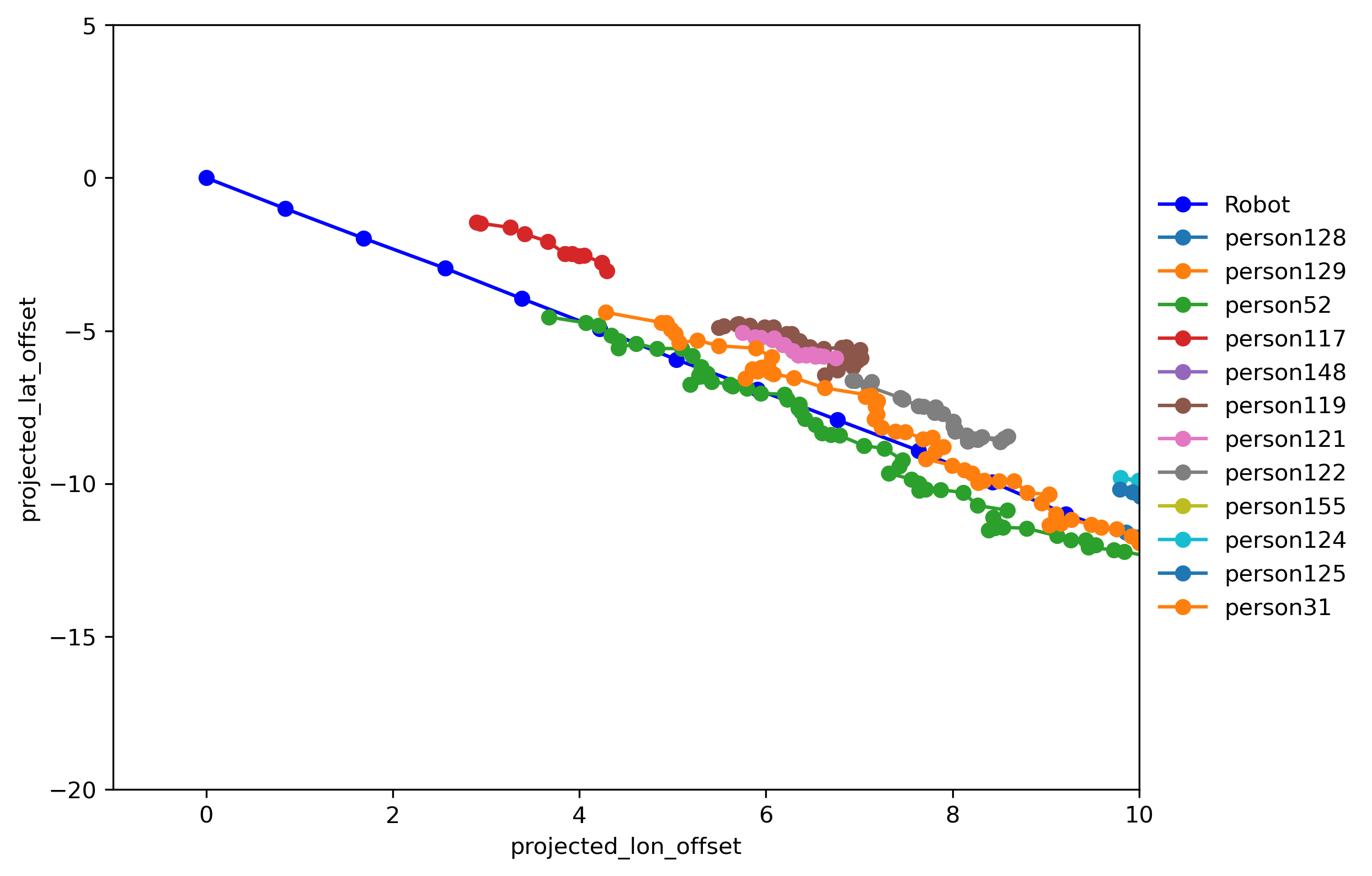}
        \caption{Pedestrian and robot trajectories}
        \label{fig:5b}
    \end{subfigure}
    \caption{Pedestrian detection}
    \label{fig:5}
\end{figure}

Pedestrians are detected in real-time using the YOLOv8 model, which assigns unique IDs to individuals (Figure~\ref{fig:5a}). Their 3D positions relative to the robot are estimated based on the depth map generated by the stereo camera. However, not all detected pedestrians are relevant for sidewalk utilization analysis. Some may be located outside the sidewalk, such as on drive lanes or opposite sidewalks, and detection accuracy decreases as the distance from the robot increases. To ensure meaningful data, only pedestrians within the sidewalk boundaries and within 10 meters of the robot are considered when computing the features.

Using the robot's GPS position, each pedestrian's absolute position can be estimated, allowing for the reconstruction of both the robot's trajectory and pedestrian trajectories, as visualized in Figure~\ref{fig:4b}. Each trajectory consists of time-stamped relative positions \((t_i,x_i,y_i)\), which are first smoothed by averaging all points within a 1-second sliding window at 0.5-second intervals to reduce noise. The average speed, speed variation, frequent turning, and path deviation are calculated for each pedestrian trajectory.

The average speed is computed by dividing the total walking distance of certain trajectory by its total travel duration. The speed variation \(\sigma_v\) is defined as the standard deviation of instantaneous speeds \(v_i\):
\begin{equation}\label{eq:3.4.1}
v_i = \frac{\sqrt{(x_{i+1} - x_i)^2 + (y_{i+1} - y_i)^2}}{t_{i+1} - t_i}
\end{equation}
The frequent turning represents the number of turning angles exceeding 30 degrees. Turning angles can be derived from successive movement bearings:
\begin{equation}\label{eq:3.4.2}
\theta_i = \arctan2(y_{i+1} - y_i, x_{i+1} - x_i)   
\end{equation}
\begin{equation}\label{eq:3.4.3}
 \Delta \theta_i = \min\left(|\theta_{i+1} - \theta_i|, 360^\circ - |\theta_{i+1} - \theta_i|\right)
\end{equation}
and we counted all \(\Delta \theta_i > 30^\circ\).

The path deviation \(\delta_{\text{path}}\) is calculated as the mean perpendicular distance from each point to the straight line connecting the start and end positions:
\begin{equation}\label{eq:3.4.4}
\delta_{\text{path}} = \frac{1}{N} \sum_{i=1}^{N} \frac{\left| (y_N - y_1) x_i - (x_N - x_1) y_i + x_N y_1 - y_N x_1 \right|}{\sqrt{(y_N - y_1)^2 + (x_N - x_1)^2}}
\end{equation}

For each sidewalk segment, the corresponding features are computed as the average of these metrics across all associated pedestrian trajectories. 

\begin{figure}[t]
    \centering
        \includegraphics[width=0.7\linewidth]{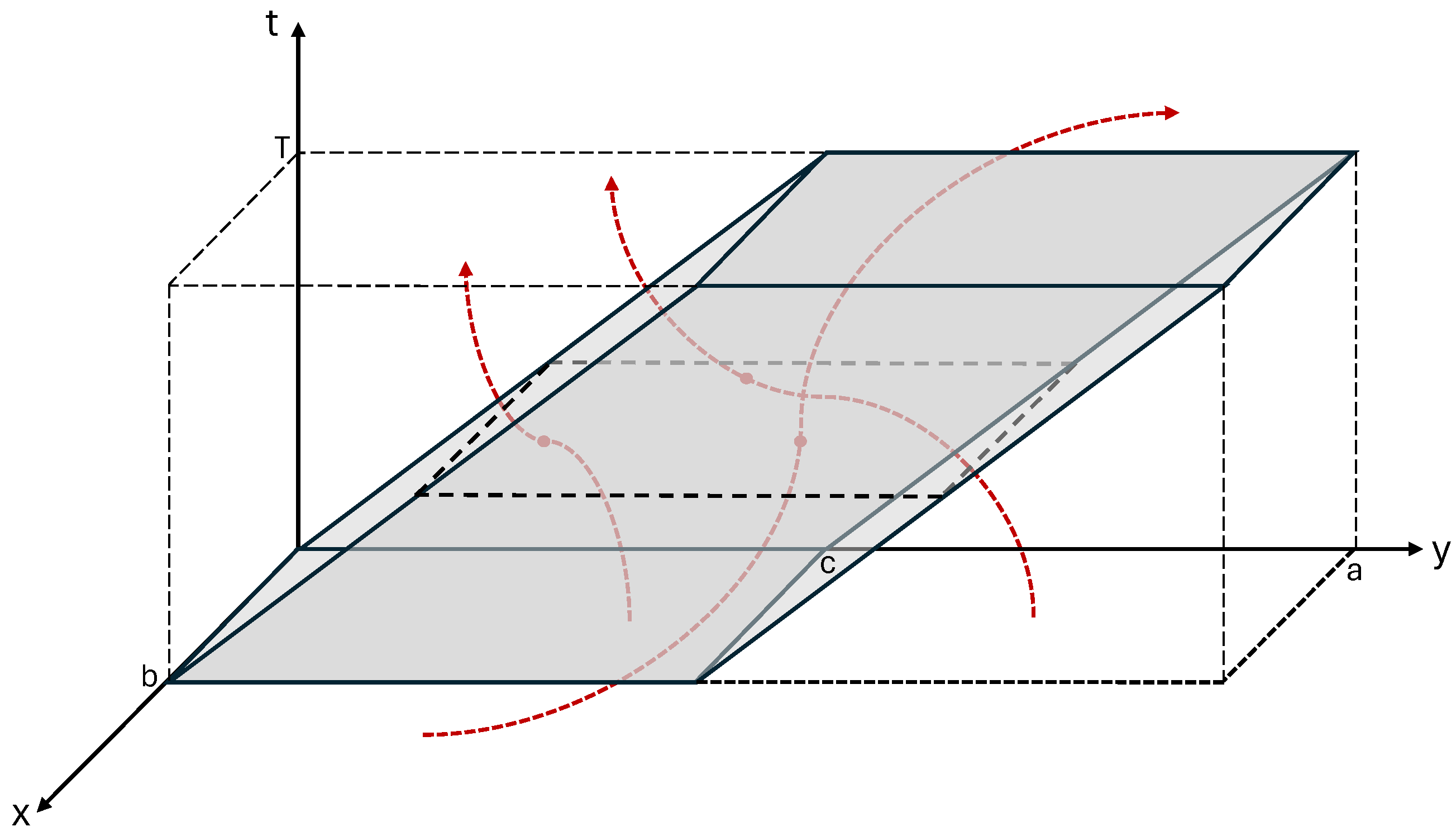}
    \caption{3D time-space diagram for pedestrian flow}
    \label{fig:6}
\end{figure}

The estimation of sidewalk pedestrian density is based on the three-dimensional pedestrian time-space diagram proposed by \citet{saberi2014exploring}, which extends Edie's definitions of fundamental traffic variables\citep{edie1963discussion}. Figure~\ref{fig:6} illustrates the three-dimensional trajectories of pedestrians measured by the robot while traversing a sidewalk segment of length \(a\) meters and width \(b\) meters, marked with red. 

The robot detects pedestrians within an effective range of \(c\) meters ahead. This range defines a detection plane of area \(|A|=bc\), which is perpendicular to the time axis. As the robot moves, this plane continuously shifts, forming a detection prism, shown in gray in Figure~\ref{fig:6}, which represents the spatiotemporal detection volume of the robot over time. Pedestrian trajectories intersect both the detection plane and the prism, allowing for density estimation under the assumption that pedestrian distribution is spatially and temporally uniform. 

At any given time \(t\), the instantaneous pedestrian density \(k(t)\) can be expressed as:
\begin{equation}\label{eq:3.4.5}
k(t) = \frac{N dt}{|A| dt} = \frac{N}{|A|}= \frac{N}{bc}
\end{equation}
where \(N\) is the number of pedestrians crossing the detection plane at time \(t\), and \(|A|=bc\) is the area of the detection plane.

Under the uniform distribution assumption, the average pedestrian density in the sidewalk segment corresponds to the average density within the detection prism, which is computed as:
\begin{equation}\label{eq:3.4.6}
k_{avg} = \frac{\sum\limits_{n \in N} \tau_n}{|V|}
= \frac{\sum\limits_{n \in N} \tau_n}{T |A|}
= \frac{\sum\limits_{n \in N} \tau_n}{T (bc)}
\end{equation}
where \(\tau_n\) is the total time spent by the pedestrian \(n\) inside the prism, \(|V|=T (bc)\) is the spatiotemporal volume of the prism, and \(T\) is the time taken by the robot to travel from the segment start to a point \(c\) meters before the segment end.

Finally, two key sidewalk utilization features are derived. The feature \texttt{max\_ped\_density} is the maximum pedestrian density \(k(t)\) observed over the duration \(T\). The feature \texttt{avg\_ped\_density} is the segment's average pedestrian density, represented by \(k_{avg}\).

\section{Results }
\label{Sec:Analysis}

We perform a series of analyses on the extracted sidewalk features to investigate their walkability-related implications. All analyses focus on sidewalks and exclude pedestrian crossings, as crossings primarily reflect intersection control, signal timing, and traffic exposure rather than sidewalk conditions or pedestrian mobility along continuous paths. Subsection \ref{sub:sidewalk-condition-feat} presents an exploratory analysis of sidewalk condition features, highlighting variation in surface quality, width, and slope across different segments. Subsection \ref{sub:sidewalk-utilization-feat} shifts the focus to aggregated pedestrian utilization patterns like speed and density relations. Building on these analyses, Subsection \ref{sub:clustering} applies clustering to identify distinct pedestrian behavior categories and analyze their relation to environmental and infrastructure characteristics. Finally, Subsection \ref{sub:regression} employs regression to investigate the potential influence of sidewalk operational features on pedestrian movement.

\subsection{Sidewalk Condition Variability Across the Network}
\label{sub:sidewalk-condition-feat}
\begin{figure}[t]
    \centering
    \begin{subfigure}{0.49\textwidth}
        \centering
        \includegraphics[width=\linewidth]{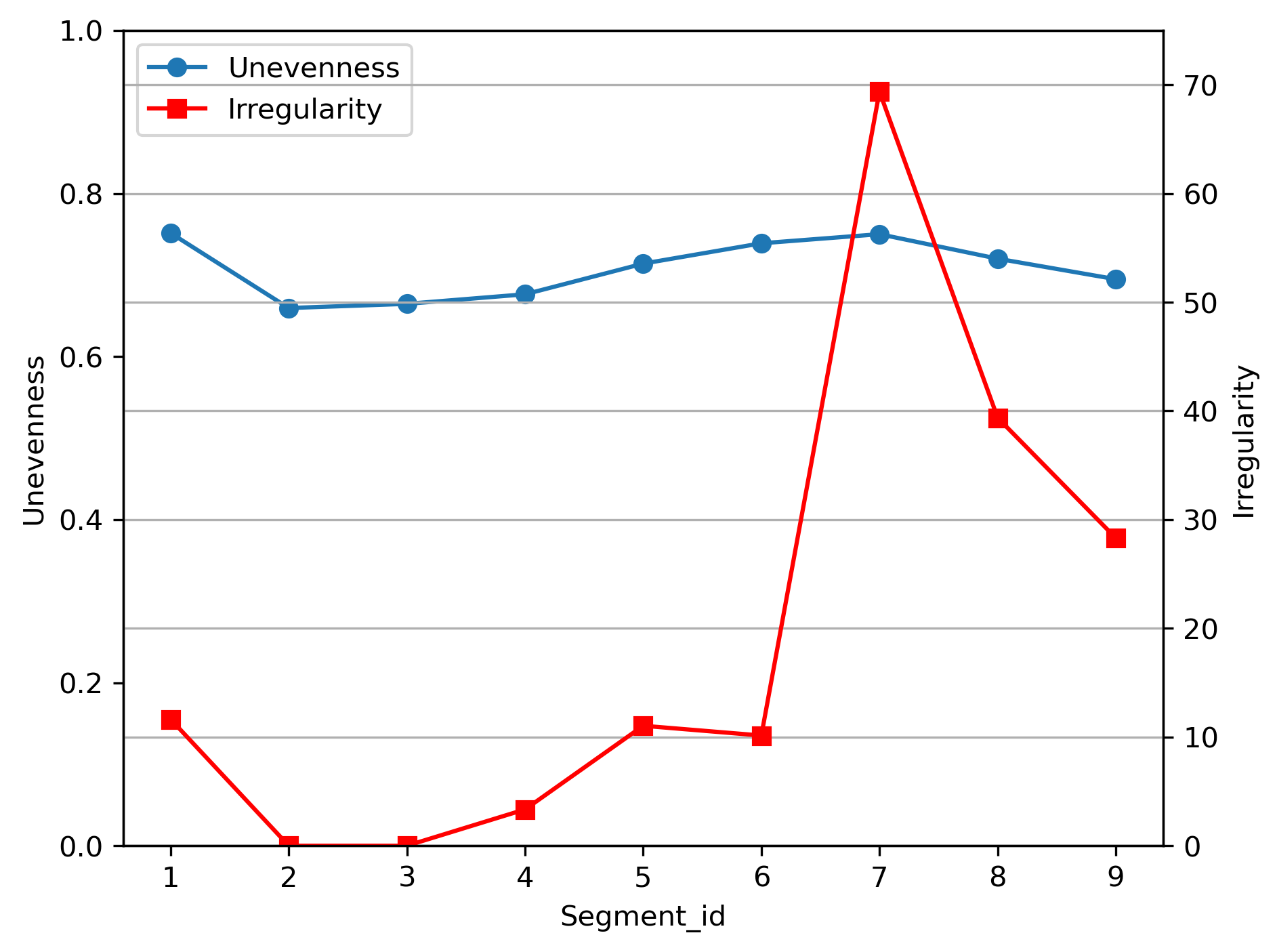}
        \caption{Indices of unevenness and irregularity across sidewalks (excluding crossings)}
        \label{fig:7a}
    \end{subfigure}
    \hfill
    \begin{subfigure}{0.49\textwidth}
        \centering
        \includegraphics[width=\linewidth]{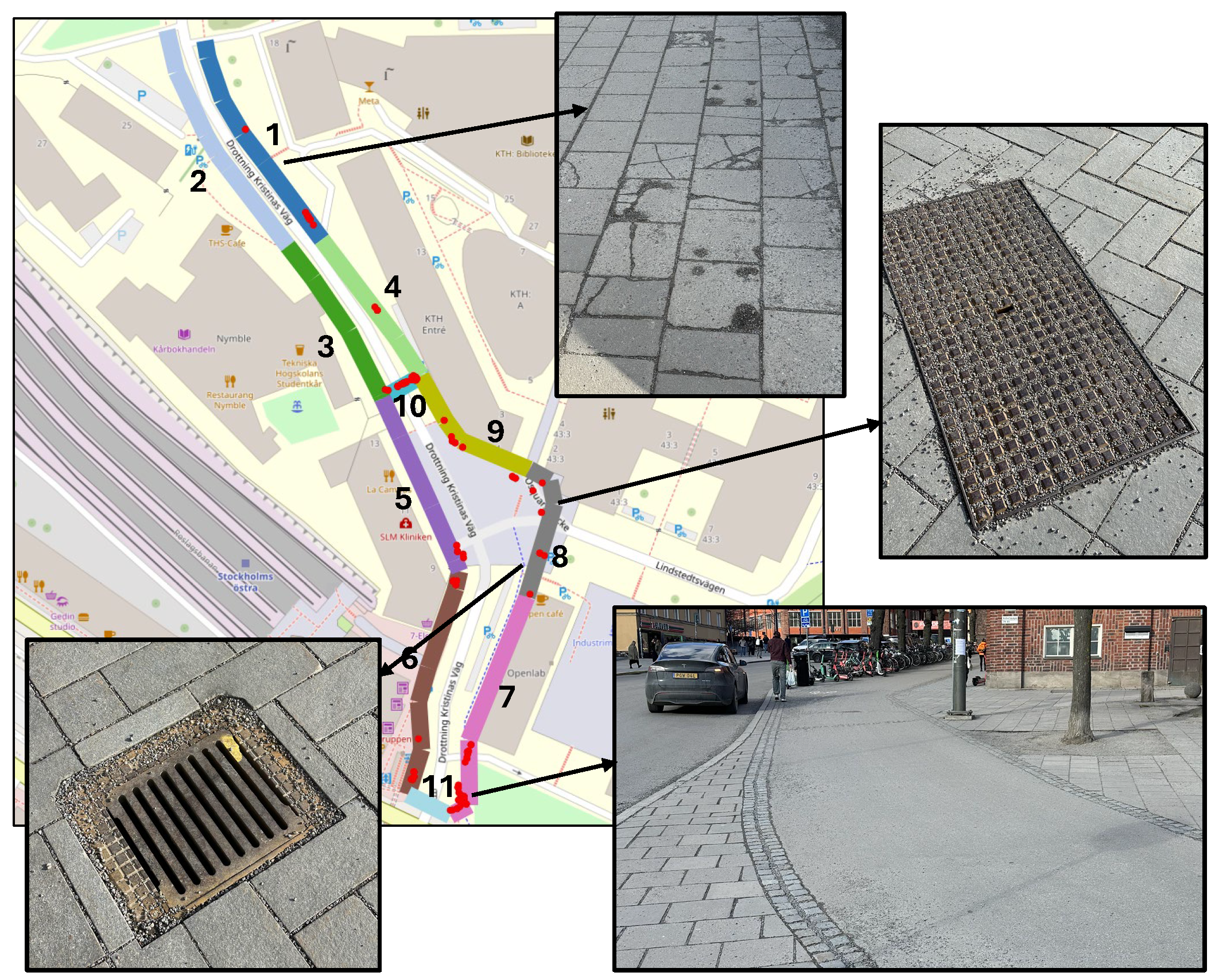}
        \caption{Examples of some critical situations on various sidewalks, with irregularity events marked as red points.}
        \label{fig:7b}
    \end{subfigure}
    \begin{subfigure}{0.49\textwidth}
        \centering
        \includegraphics[width=\linewidth]{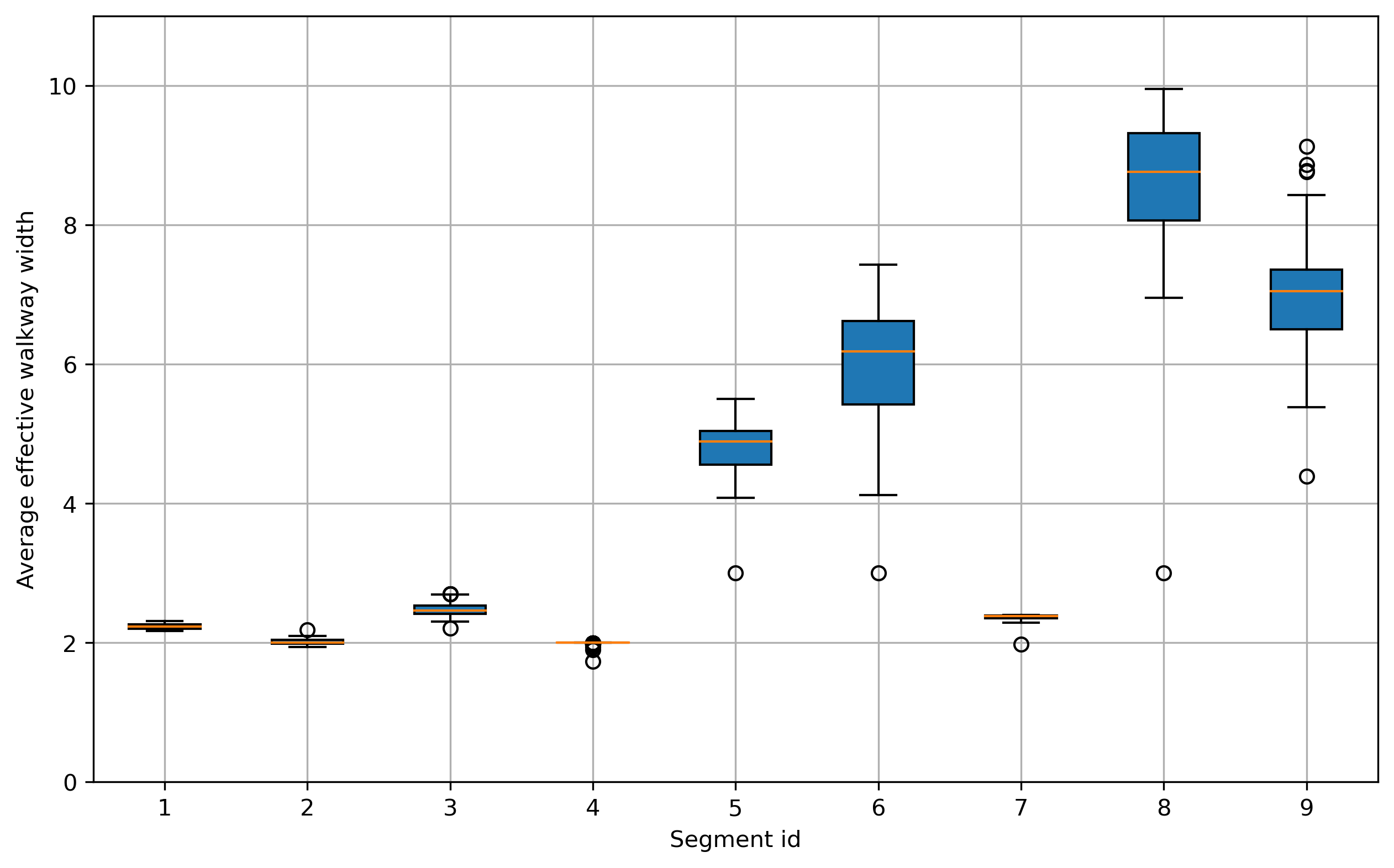}
        \caption{Average effective width across sidewalks}
        \label{fig:7c}
    \end{subfigure}
    \hfill
    \begin{subfigure}{0.49\textwidth}
        \centering
        \includegraphics[width=\linewidth]{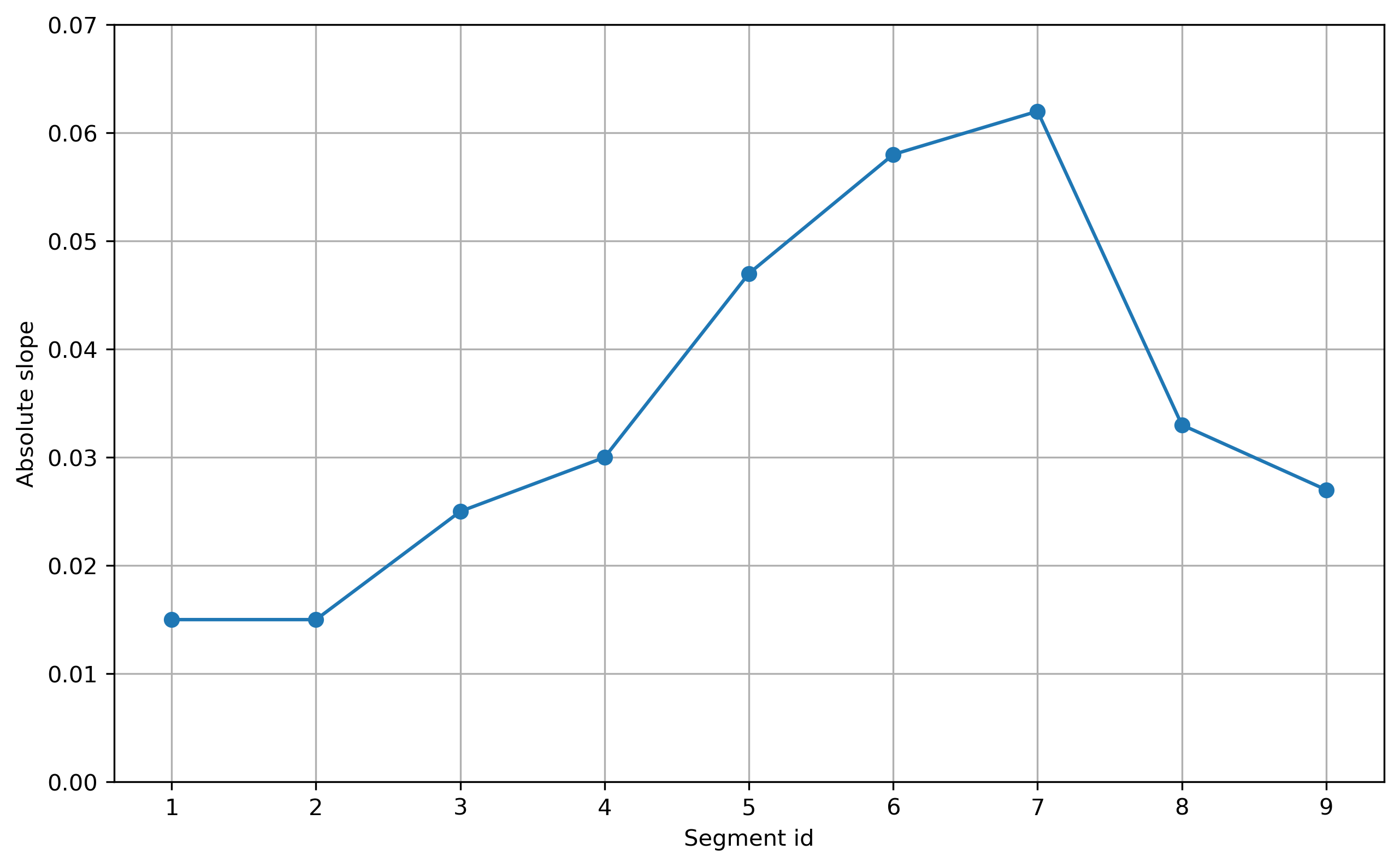}
        \caption{Absolute slope across sidewalks}
        \label{fig:7d}
    \end{subfigure}
    \caption{Comparisons of sidewalk condition features across multiple sidewalks}
    \label{fig:7}
\end{figure}

Indices of unevenness and irregularity across nine sidewalk segments are illustrated in Figure~\ref{fig:7a}. The unevenness index is relatively consistent across segments, generally averaging around 0.7, indicating comparable surface conditions in terms of gradual variations. Segment 1 shows the highest unevenness, primarily due to the presence of broken and uneven bricks (see Figure~\ref{fig:7b}). 

In contrast, the irregularity index exhibits higher variability across the network, with Segment 7 displaying a significant spike due to frequent paving pattern changes (see Figure~\ref{fig:7b}). When the robot travels through this segment, it encounters four abrupt paving transitions within a short distance, including a cycling lane in between. Segment 7 records the highest irregularity but only moderate unevenness, indicating that the two IMU-based metrics capture complementary aspects. Irregularity is also elevated on Segment 8 (manhole cover, storm drain), whereas Segments 2–3 are minimal. Together, these results confirm that our proposed unevenness and irregularity metrics accurately describe actual sidewalk conditions, providing valuable insights for assessing pedestrian comfort and sidewalk robot mobility.   

The distributions of detected average effective walkway width during the robot's trips across various sidewalks are shown in Figure~\ref{fig:7c}. Narrow sidewalks (Segments 1-4 and 7) display tight width distributions, generally below 3 m, while open areas (around 5.5–10 m) lack clear edges and show greater variability. This variation reflects both geometric differences and the robot’s sensing limitation. The stereo camera which offers a 120-degree field of view, in practice captures sidewalk information up to a maximum of about 7 meters ahead when centered. In broader or less structured spaces, the robot’s steering to avoid pedestrians or obstacles can further underestimate or fluctuate measured width. Consequently, width estimates are most reliable in narrow, well-defined corridors, while some variation in these areas may still result from temporary obstacles encountered during passage. 

Figure~\ref{fig:7d} presents the absolute slope values for each sidewalk segment, capturing steepness differences relevant to walkability and robot navigation. Segment 7 is the steepest, whereas Segments 1 and 2 are nearly flat, reflecting modest but notable variations across the network.  

Overall, surface quality, effective width, and slope show clear segment-level variability that shapes the walking environment and provides the basis for the utilization and behavioral analyses in subsequent sections.

\subsection{Sidewalk Utilization and Density-Speed Dynamics}
\label{sub:sidewalk-utilization-feat}

\begin{figure}
    \centering
    \begin{subfigure}{0.49\textwidth}
        \centering
        \includegraphics[width=\linewidth]{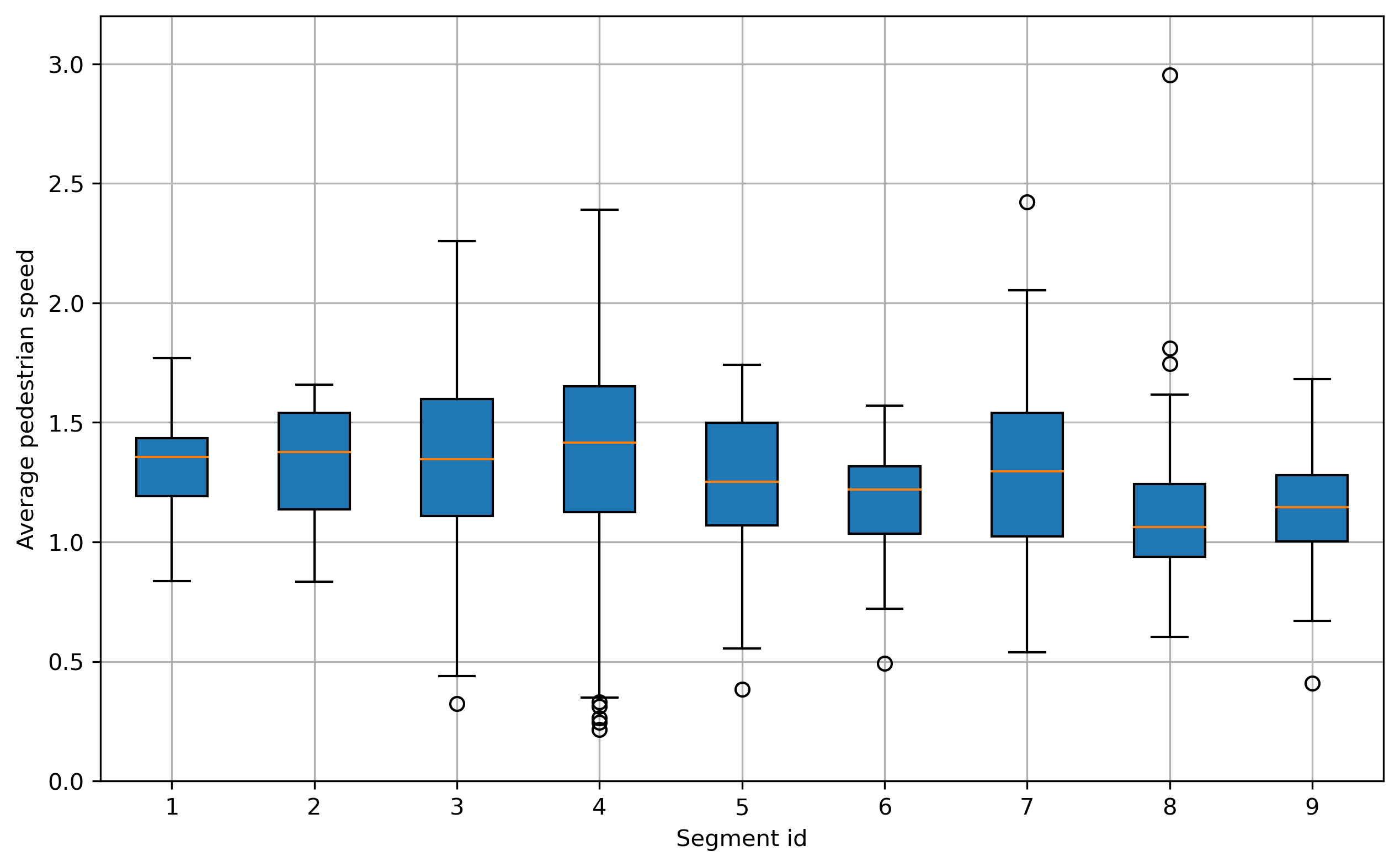}
        \caption{Average pedestrian speed across sidewalks}
        \label{fig:8a}
    \end{subfigure}
    \hfill
    \begin{subfigure}{0.49\textwidth}
        \centering
        \includegraphics[width=\linewidth]{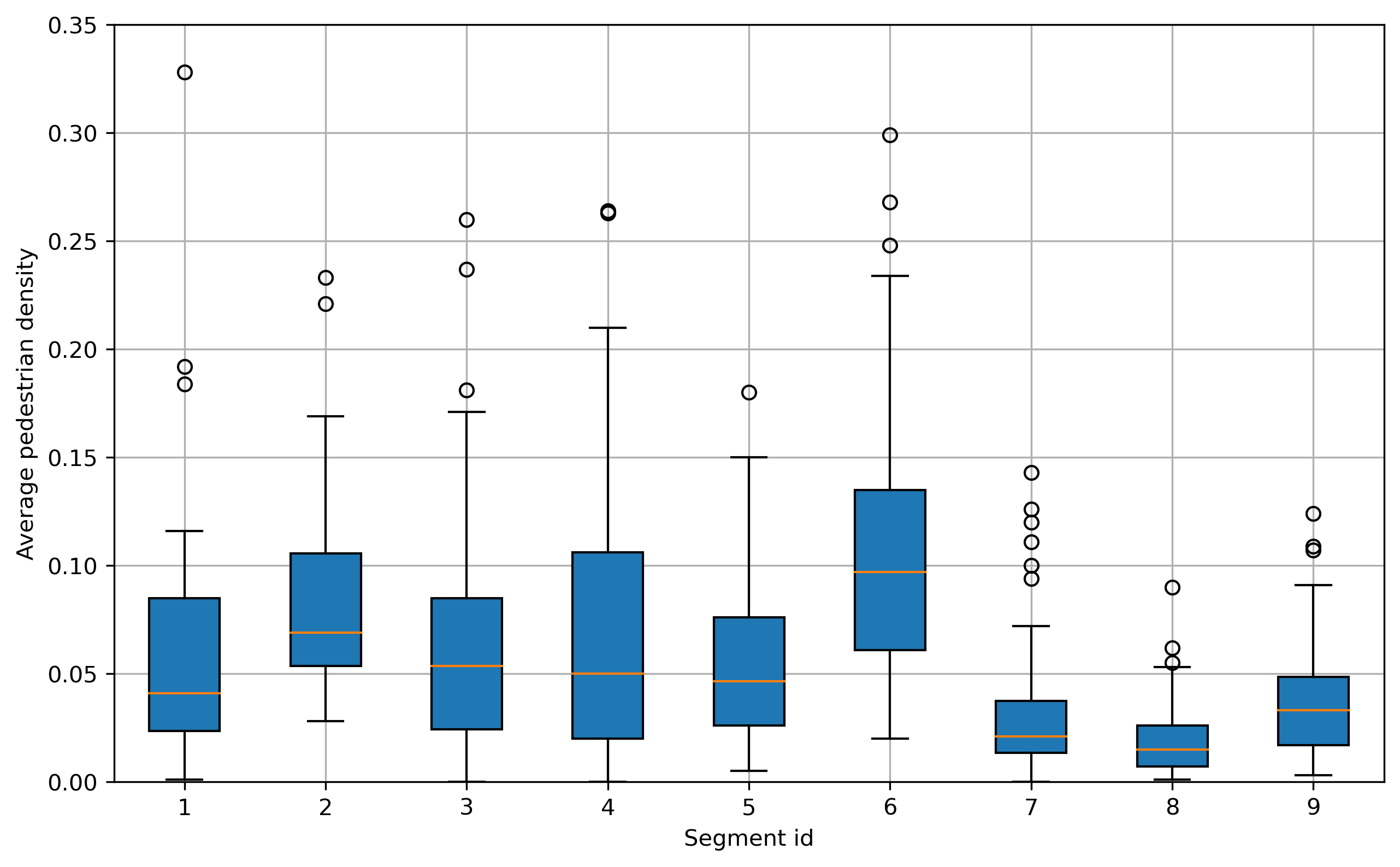}
        \caption{Average pedestrian density across sidewalks}
        \label{fig:8b}
    \end{subfigure}
    \caption{Comparisons of sidewalk utilization features across multiple sidewalks}
    \label{fig:8}
\end{figure}

Average pedestrian speed and pedestrian density are estimated during each robot traversal of the sidewalk segments. The distributions of these two features across different sidewalks are presented in Figure~\ref{fig:8a} and Figure~\ref{fig:8b}. The results reveal notable variability in pedestrian behavior across segments. Segments 3 and 4 show broader speed distributions, suggesting a mix of movement patterns and variable flow conditions, while Segments 6 and 9 display more concentrated speeds, indicating steadier movement.

Pedestrian density also differs across sidewalks. Although Segments 6 and 7 are located on opposite sides of the same road, Segment 6 shows the highest density, while Segment 7 remains sparse. As shown in Figure~\ref{fig:7c}, their average sidewalk widths are approximately 6 m and 2 m, respectively, illustrating how available space influences pedestrian volumes. Segment 6 serves as the main access corridor connecting the metro and light rail station exits to the campus entrance, and it is lined with convenience stores and restaurants. By contrast, segment 7 lacks notable points of interest, resulting in limited pedestrian activity.  

\begin{figure}
    \centering
    \begin{subfigure}{0.48\textwidth}
        \centering
        \includegraphics[width=\linewidth]{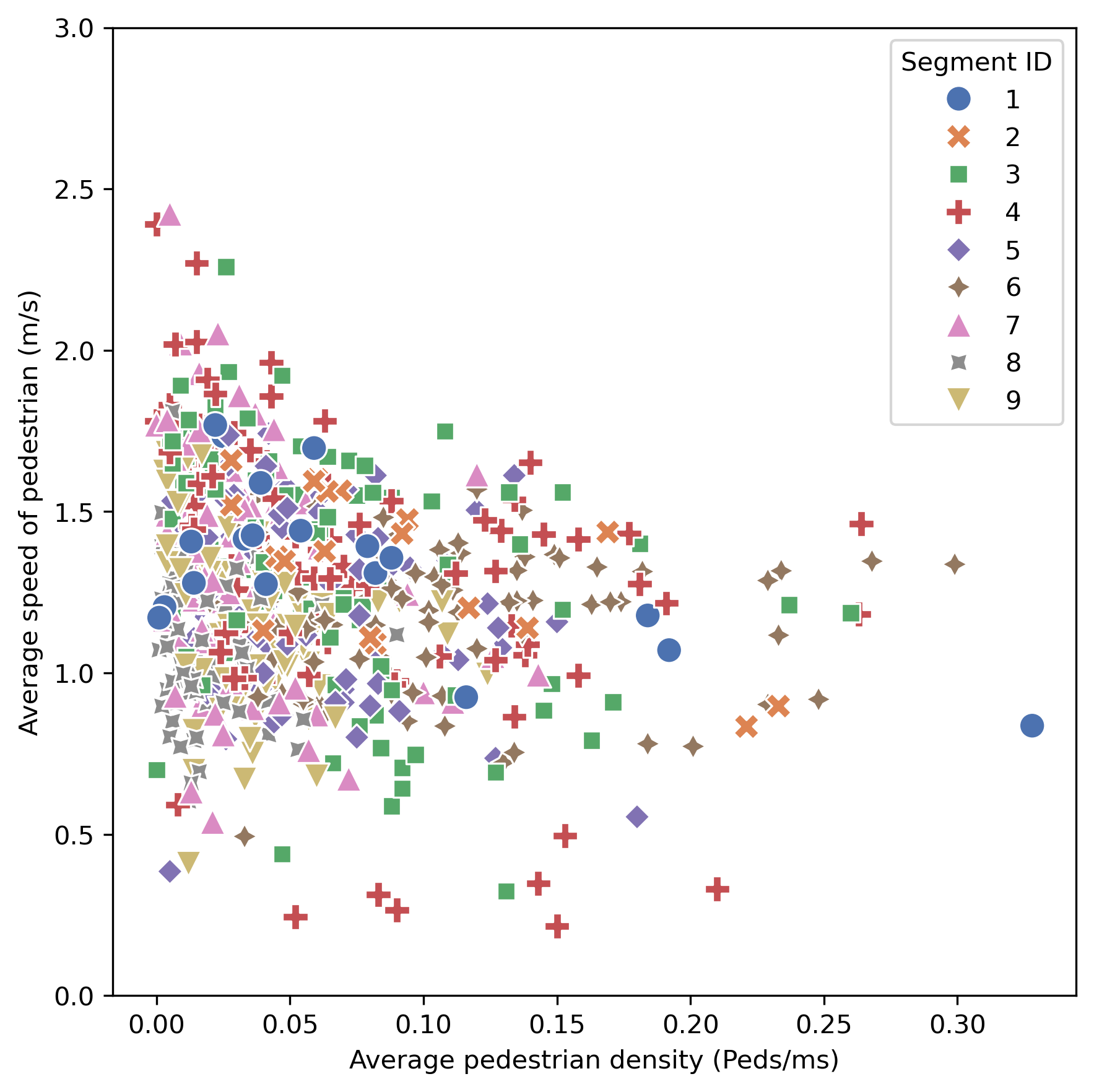}
        \caption{Speed density relation in this study}
        \label{fig:9a}
    \end{subfigure}
    \hfill
    \begin{subfigure}{0.50\textwidth}
        \centering
        \includegraphics[width=\linewidth]{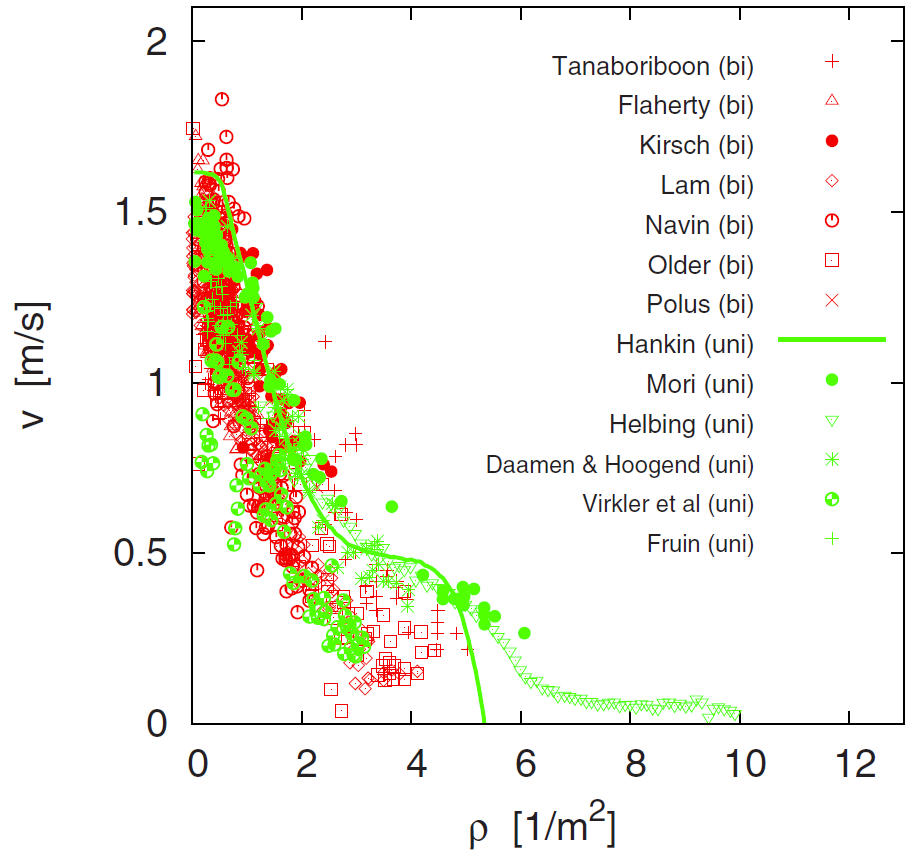}
        \caption{Speed density relation in previous studies \citep{zhang2013empirical}}
        \label{fig:9b}
    \end{subfigure}
    \caption{Density - speed fundamental diagrams of pedestrian flow from this study and previous studies}
    \label{fig:9}
\end{figure}

Figure~\ref{fig:9a} presents the fundamental diagram of bidirectional pedestrian flow on sidewalks, which illustrates the relationship between average pedestrian density and average pedestrian speed. The previous studies on pedestrian flow dynamics have primarily focused on indoor environments, such as corridors \citep{zhang2012high}, bottlenecks \citep{tian2012experimental}, stairs \citep{burghardt2013performance}, and escalators \citep{lee2005pedestrian}. Unlike indoor facilities, outdoor sidewalks often lack strict physical boundaries. In the studied network, sidewalks are flanked by grass and driving lanes, allowing pedestrians to dynamically adjust their walking paths. It is frequently observed that pedestrians walk on adjacent grass or even step into driving lanes to avoid congestion or maintain their pace, rather than slowing down or waiting. For comparison, the fundamental diagram of pedestrian flow derived from prior empirical studies, as summarized by \citet{zhang2013empirical}, is provided in Figure~\ref{fig:9b}.

The observed densities in our dataset are relatively low ( up to 0.34~peds/m\textsuperscript{2}), while walking speeds reach 2.5 m/s, reflecting the open campus setting where pedestrians move purposefully in an outdoor environment. Despite these scale differences, the overall pattern follows the expected inverse relationship between density and speed: higher densities correspond to slower movement. Measurement errors due to the robot’s onboard hardware and perception-limited capabilities may also contribute to such discrepancy. The robot relies on real-time pedestrian-detection algorithms \citep{ultralytics2023yolov8}, which are constrained by onboard computational power and may not achieve the precision of offline or stationary sensing systems. 

At low densities (below 0.1~peds/m\textsuperscript{2}), speeds vary widely from under 0.5~m/s to over 2~m/s, reflecting heterogeneous walking behaviors. As density increases beyond 0.15~peds/m\textsuperscript{2}, speeds converge below 1.5~m/s, suggesting the onset of congestion and greater interaction among pedestrians. Segment-level variations also emerge. For instance, Segment 4 exhibits a broader spread of speeds, while Segment 1 consistently supports higher walking speeds even at moderate densities. These patterns suggest that both spatial context and sidewalk characteristics influence how pedestrians adapt their pace in shared environments. 

The analysis confirms the expected inverse relationship between pedestrian density and speed, while also revealing context-specific variations across sidewalks. Together, these findings characterize how pedestrians collectively respond to spatial constraints and flow conditions, offering empirical grounding for the behavioral analyses that follow.

\subsection{Pedestrian Behavior Patterns and Sidewalk Features}

\subsubsection {Behavioral Clustering}
\label{sub:clustering}
To explore whether specific sidewalk features are associated with variations in pedestrian movement patterns, we implement a clustering-based analysis using high-resolution trajectory data collected by the robot.  The approach relies on hierarchical agglomerative clustering applied to sidewalks, based on three pedestrian behavior features: pedestrian speed variation, turning frequency, and path deviation. Outliers in turning behavior using the interquartile range (IQR) method are removed before clustering to ensure that extreme turning events due to noise and anomalous events are not considered. 

\begin{figure}[ht]
    \centering
        \includegraphics[width=1\linewidth]{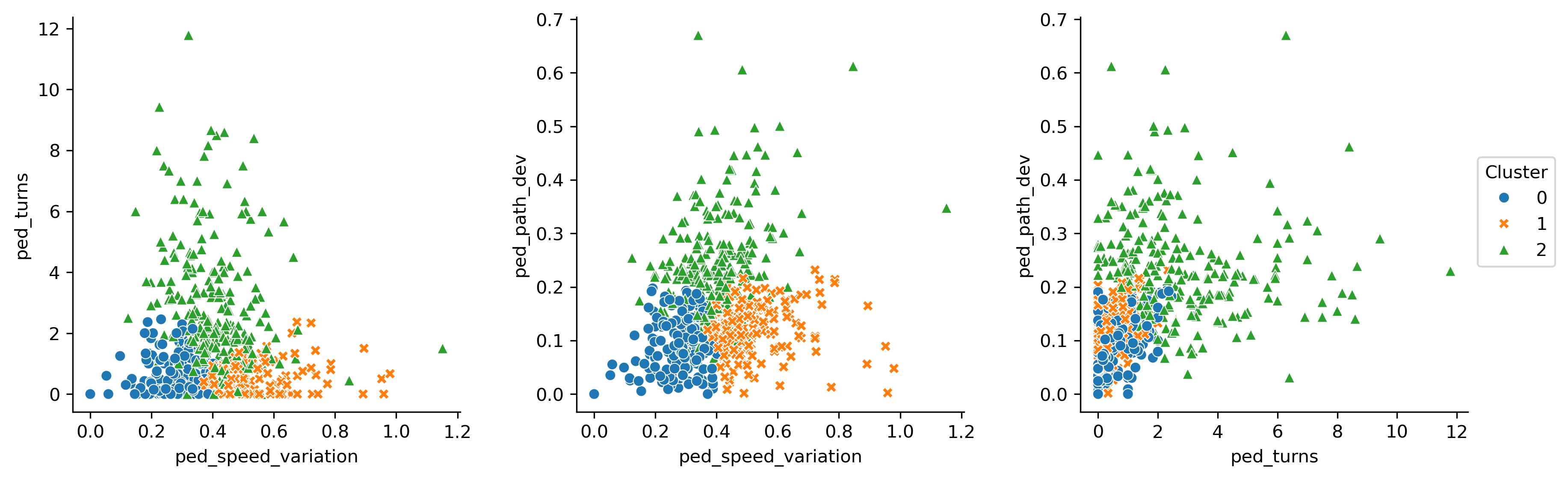}
    \caption{Hierarchical Clustering of Pedestrian Behavior: Pedestrian speed variation, turning frequency, and path deviation}
    \label{fig:10}
\end{figure}

The number of clusters was determined through inspection of the resulting dendrogram, which revealed a natural break at three clusters. This selection was further validated by examining the separability of groups in pairwise scatter plots, where distinct behavioral patterns became evident across all three dimensions. The final clustering solution revealed three meaningful behavior groups, as shown in Figure~\ref{fig:10}. 

Each cluster reflects a distinct pedestrian behavioral pattern. Cluster 0 comprises 169 records and is characterized by relatively lower pedestrian speed variation, infrequent turning, and minimal path deviation. These patterns suggest a group of stable, efficient walkers who traverse sidewalks in a direct and predictable manner, likely with clear destinations and minimal distractions. Cluster 1 contains 183 records and captures dynamic, straight-line behavior, with individuals exhibiting the highest speed variation, fewest turns, and moderate path deviation. Cluster 2 (green) includes 244 records and is defined by the highest turning frequency and the most significant path deviation, while maintaining relatively moderate speed variation.

\begin{table}[t]
\centering
\caption{Descriptive statistics of pedestrian behavior and sidewalk features by cluster}
\label{tab:3}
\resizebox{\textwidth}{!}{%
\begin{tabular}{llll}
\toprule
\textbf{Feature} & \textbf{Cluster 0 (n = 169)}& \textbf{Cluster 1 (n = 183)}& \textbf{Cluster 2 (n = 244)}\\
\midrule
ped\_speed\_variation (m/s)& 0.266 ± 0.088      & 0.514 ± 0.117***   & 0.405 ± 0.112***      \\ 
ped\_turns            & 0.629 ± 0.642      & 0.456 ± 0.463**   & 2.815 ± 2.052***   \\ 
ped\_path\_deviation(m)& 0.089 ± 0.055      & 0.122 ± 0.048***   & 0.240 ± 0.099***   \\ 
slope                 & -0.001 ± 0.038     & 0.002 ± 0.043     & -0.003 ± 0.040   \\ 
avg\_effective\_width (m)& 4.983 ± 2.596      & 3.937 ± 2.194***    & 4.704 ± 2.492   \\ 
irregularity\_index   & 25.093 ± 21.527    & 22.593 ± 25.354    & 15.619 ± 18.081***   \\ 
unevenness\_index     & 0.707 ± 0.028      & 0.709 ± 0.033   & 0.702 ± 0.030   \\ 
avg\_ped\_density (Peds/m²)& 0.034 ± 0.035      & 0.048 ± 0.047**      & 0.072 ± 0.056***   \\ 
avg\_ped\_speed (m/s)& 1.254 ± 0.377      & 1.451 ± 0.242***      & 1.168 ± 0.255*   \\ 
\bottomrule
\end{tabular}
}
\vspace{0.2cm}
\footnotesize
Significance is tested against Cluster 0. 
*p $<$ 0.05, **p $<$ 0.01, ***p $<$ 0.001.
\end{table}

To investigate the environmental correlates of these behaviors, we examined both sidewalk condition and utilization characteristics across clusters, as shown in Table~\ref{tab:3}. For each feature, the mean and standard deviation were calculated per cluster to capture both central tendency and variability.  Given that Cluster 0 reflects typical, stable walking behavior, it serves as the reference group. Statistical significance of differences in Cluster 1 and Cluster 2 relative to Cluster 0 was assessed using independent t-tests, with significance levels indicated by asterisks. 

The comparison of sidewalk condition and utilization features across behavioral clusters reveals different patterns. Cluster 0, which is indicative of stable and direct walking behavior, is associated with the widest average effective sidewalk width (4.983 m) and lowest pedestrian density (0.034 Peds/m²). This suggests that stable, consistent walking tends to occur in more spacious environments with freedom of movement. 

Cluster 1, characterized by high speed variation but the fewest turns, is linked to narrower sidewalks (average effective width: 3.937 m) and moderate crowding (0.048 peds/m²). These relatively constrained spatial conditions may force pedestrians to frequently adjust their speed or momentarily pause, particularly when navigating around other users or avoiding obstacles, leading to the erratic behavior observed. Cluster 1 also exhibits the highest walking speed (1.451 m/s) across all clusters, possibly reflecting an effort to move quickly through constrained spaces. 

In contrast, Cluster 2 reflects a behavior characterized by moderate speed variation, frequent turning, and higher path deviation. Interestingly, this cluster is associated with the highest pedestrian density (0.072 peds/m²), the lowest irregularity index (15.62), and shows the lowest average walking speed (1.168 m/s) among the three groups. While these patterns suggest more complex or constrained movement, particularly in denser settings, further investigation would be required to interpret their underlying causes.

No statistically significant differences were found in slope or unevenness index across the clusters, indicating that these surface features may play a lesser role in shaping observed behavioral differences within the current sample.

These findings support the notion that pedestrian movement behavior is not only shaped by individual intent but also by a dynamic interplay with the sidewalk environment. Wider and less crowded sidewalks appear to support stable and direct walking, while narrower and denser environments can lead to more variable and hesitant movement patterns. Smoother environments, especially under crowded conditions, may inadvertently encourage non-linear movement. A more robust understanding requires further analysis across a larger urban network and with a more extensive dataset, in order to validate these relationships and assess their broader applicability.

\subsubsection {Effects of Sidewalk Features on Pedestrian Speed}
\label{sub:regression}

\begin{figure}[ht]
    \centering
        \includegraphics[width=0.9\linewidth]{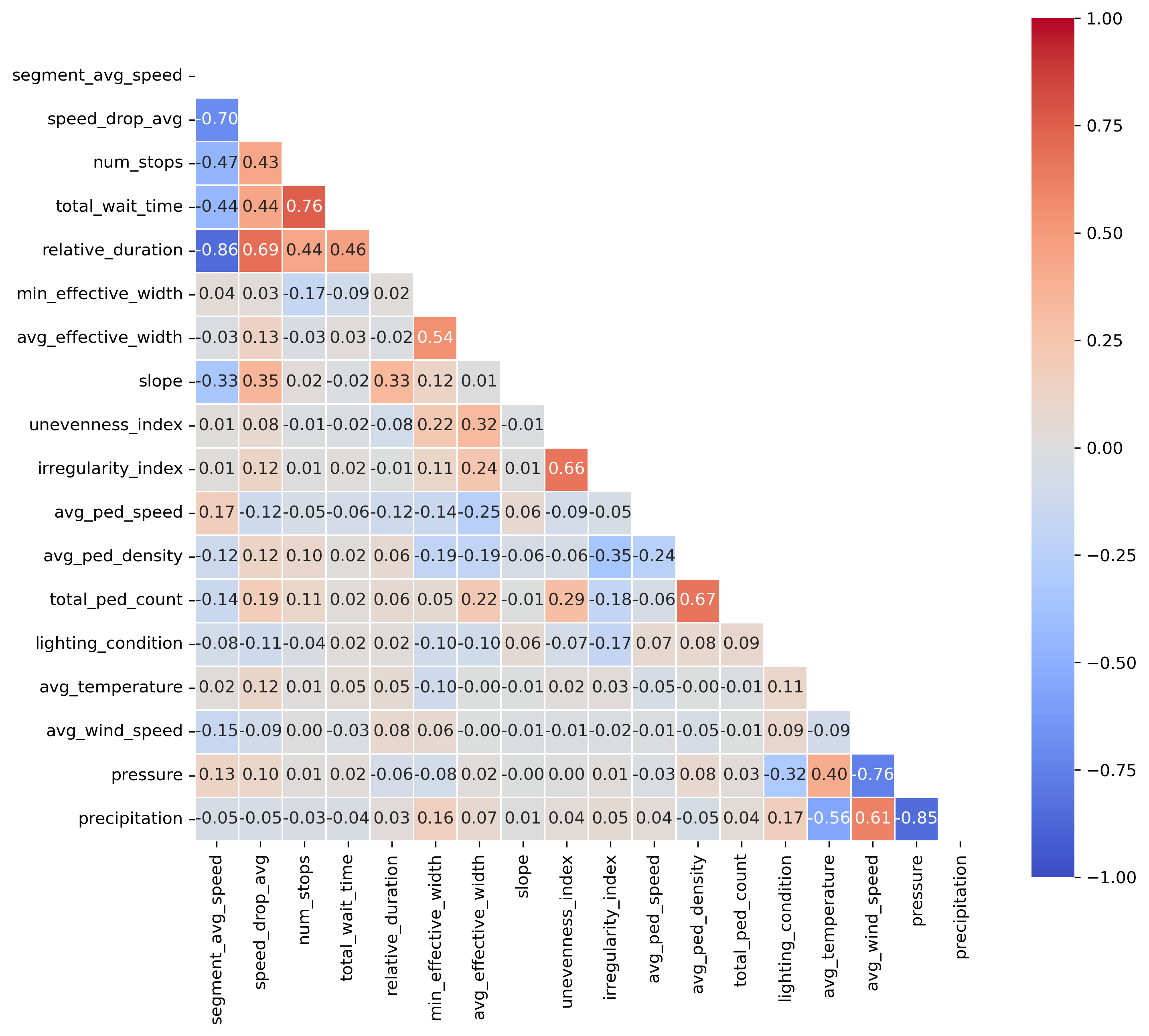}
    \caption{Correlation matrix of robot trip features and sidewalk features for all sidewalks}
    \label{fig:11}
\end{figure}

To complement the cluster-based findings presented in the previous section, we investigate how aggregated measures, such as pedestrian density, relate to sidewalk characteristics and robot operational performance. Specifically, pairwise Pearson correlation coefficients were computed across key robot trip features, sidewalk condition features, and sidewalk utilization features. In addition to features collected from robots, weather-related features, such as temperature, wind speed, pressure, and precipitation, were incorporated from external data sources to assess their potential influence. The resulting correlation matrix is shown in Figure~\ref{fig:11}.

Strong negative correlations are observed between the robot's average speed and several trip-related delay metrics. Notably, relative duration, which has the strongest negative correlation with robot average speed ($-0.86$), underscores its sensitivity to disturbances during robot navigation.

Among sidewalk condition features, minimum effective width exhibits weak negative correlations with the number of robot stops ($-0.17$), but shows almost no relationship with the robot's average speed ($-0.04$). This indicates that the robots are more likely to stop in narrower areas, such as sidewalk bottlenecks, but these brief interruptions do not significantly affect the overall average speed. In contrast, average effective width is negatively correlated with average pedestrian speed ($-0.25$), suggesting that pedestrians may walk faster on generally narrower sidewalks.

Slope is weakly negatively correlated with robot average speed ($-0.33$), and positively correlated with both speed drop average ($0.35$) and relative duration ($0.33$), indicating that slopes can slow robot movement and contribute to travel time variability. However, slope has minimal correlation with average pedestrian density ($-0.06$) and average pedestrian speed ($-0.06$). This suggests that although slope affects robot performance, it does not significantly influence pedestrian flow in this context. 

Average pedestrian speed shows weak negative correlations with average pedestrian density ($-0.24$), and a weak positive correlation with robot average speed ($0.17$). The inverse relationship with density reflects expected pedestrian dynamics, where higher densities typically slow down movement. Similarly, the positive association with robot average speed is also consistent with expectations, as robots are likely to operate faster in less congested environments. Additionally, average pedestrian density shows a weak positive correlation with speed drop average ($0.12$), suggesting that increased crowding can contribute to momentary slowdowns in robot travel.

Weather variables do not exhibit strong correlations with sidewalk or robot performance features in this dataset, but they may become more relevant in long-term or seasonal analyses and should be considered in future research. 

To further examine how various sidewalk environmental and robot operational factors influence pedestrian movement, we perform a multiple linear regression using average pedestrian speed as the dependent variable. The initial full model includes 15 features, covering sidewalk condition features, pedestrian utilization features, and robot trip features (Table~\ref{tab:4}). To address potential multi-collinearity and ensure the interpretability of regression coefficients, we selectively excluded highly correlated variables based on the correlation matrix (Figure~\ref{fig:11}), retaining only a subset of less correlated variables for the final regression analysis. All input features were normalized to a [0,1] range. Additionally, average pedestrian density was log-transformed to reduce skewness and improve linearity in its relationships with the dependent variable. 

\begin{table}[t]
\centering
\begin{threeparttable}
\caption{Regression Results of the full model and reduced model explaining Average Pedestrian Speed}
\label{tab:4}
\begin{tabular}{lcclccl}
\toprule
\multirow{2}{*}{Variable} & \multicolumn{3}{c}{Full Model} & \multicolumn{3}{c}{Reduced Model} \\
\cmidrule(lr){2-4} \cmidrule(lr){5-7}
& Estimate & Std. Error & P-value & Estimate & Std. Error & P-value \\
\midrule
(Intercept) & 0.774& 0.252& 0.002 $^{**}$& 0.894& 0.150& 0.000 $^{***}$\\
avg\_ped\_density & -0.732& 0.092& 0.000 $^{***}$& -0.686& 0.089& 0.000 $^{***}$\\
segment\_avg\_speed & 0.707& 0.255& 0.006 $^{**}$& 0.695& 0.156& 0.000 $^{***}$\\
speed\_drop\_avg & 0.382& 0.171& 0.026 $^{*}$& 0.398& 0.155& 0.010 $^{*}$\\
num\_stops & 0.068& 0.123& 0.582& -- & -- & -- \\
relative\_duration & -0.022& 0.284& 0.939& -- & -- & -- \\
min\_effective\_width & -0.190& 0.116& 0.102& -- & -- & -- \\
avg\_effective\_width & -0.316& 0.056& 0.000 $^{***}$& -0.353& 0.048& 0.000 $^{***}$\\
slope & 0.098& 0.047& 0.039 $^{*}$& 0.085& 0.046& 0.065 ${.}$\\
unevenness\_index & 0.131& 0.059& 0.028 $^{*}$& 0.112& 0.058& 0.054 ${.}$\\
irregularity\_index & -0.200& 0.067& 0.003 $^{**}$& -0.178& 0.065& 0.006 $^{**}$\\
 lighting\_condition & 0.176& 0.083& 0.034 $^{*}$& 0.132& 0.069&0.057 ${.}$\\
avg\_temperature& -0.135& 0.079& 0.090 ${.}$& -0.112& 0.053& 0.037 $^{*}$\\
avg\_wind\_speed& 0.078& 0.080& 0.333& -- & -- & -- \\
pressure& 0.180& 0.113& 0.113& -- & -- & -- \\
precipitation & 0.095& 0.089& 0.291& -- & -- & -- \\
\midrule
\multicolumn{1}{l}{} & \multicolumn{3}{l}{R$^2$: 0.217, Adj. R$^2$: 0.194} & \multicolumn{3}{l}{R$^2$: 0.208, Adj. R$^2$: 0.194} \\
\multicolumn{1}{l}{} & \multicolumn{3}{l}{Obs: 516, F: 9.261, p $<$ 0.001} & \multicolumn{3}{l}{Obs: 516, F: 14.76, p $<$ 0.001} \\
\bottomrule
\end{tabular}
\begin{tablenotes}
\small
\item ${.}$p $<$ 0.1, $^{*}$p $<$ 0.05, $^{**}$p $<$ 0.01, $^{***}$p $<$ 0.001. “--” indicates variable not included in model.
\end{tablenotes}
\end{threeparttable}
\end{table}

The full model reveals that several features are significantly associated with pedestrian speed. The average pedestrian density and average effective width exhibit strong negative influences ($p<0.001$ for both), while the segment average speed shows strong positive effects ( $p=0.006$). However, several features, such as the number of stops, relative duration, unevenness index, average wind speed, and precipitation, yield high p-values and contribute little explanatory power. The full model achieves an adjusted $R^2$ of 0.194.

A reduced model was constructed to refine the model and improve robustness, by removing variables with p-values higher than \(0.1\). The removed features include the number of stops, relative duration, minimum effective width, average wind speed, pressure, and precipitation. The resulting reduced model achieves nearly the same performance (with adjusted  $R^2$ of 0.194) with fewer variables, and higher F-statistics (14.76 vs 9.261), suggesting that the remaining variables are more statistically meaningful and contribute more effectively to explaining the variance in the average pedestrian speed.

In the reduced model, average pedestrian density remains a strong negative predictor ($\beta = -0.686$, $p<0.001$), consistent with the expected crowding effect. Average effective width also exhibits a significant negative relationship with pedestrian speed  ($\beta = -0.353$, $p<0.001$), consistent with the earlier correlation result, indicating that pedestrians tend to walk more slowly on wider sidewalks and faster on narrower ones. Segment average speed emerges as the most influential factor  ($\beta = 0.695$, $p<0.001$), suggesting that robot speed performance closely reflects pedestrian movement and may serve as a useful feature for estimating walkability.

Additionally, the irregularity index shows a highly significant negative effect on pedestrian speed ($\beta = -0.178$, $p=0.006$), implying that poor surface conditions may discourage faster walking. Daily average temperature is also a statistically significant negative predictor  ($\beta = -0.112$, $p=0.037$), showing that colder weather may lead to pedestrians walking more briskly. 

Interestingly, average speed drop shows a statistically positive relationship with pedestrian speed ($\beta = 0.398$, $p=0.01$). This seems counterintuitive since more robot speed drops may suggest poor walkability due to congestion or uneven terrain. However, this appears to reflect short, sharp avoidance maneuvers occurring in otherwise free-flow conditions. Because robots are less agile than pedestrians and adopt more conservative avoidance strategies, these maneuvers register as speed drops even in segments where pedestrians maintain relatively high walking speeds.

Overall, the reduced model presents a set of interpretable and meaningful features that influence pedestrian speed in urban sidewalk environments. These findings offer valuable insights for urban planners assessing walkability in the future.

\section{Discussion}
\label{Sec:discussion}

Our findings both confirm and add new insights to the established knowledge of pedestrian dynamics. The clustering results reveal that pedestrian movement behavior is shaped not only by individual intent but also by continuous interaction with the surrounding sidewalk environment. Specifically, wider and less crowded sidewalks tend to support more stable and direct walking, whereas narrower and denser environments can lead to more variable and hesitant movement patterns. This observation echoes prior work linking sidewalk width and buffers to perceived safety and smoother walking \citet{su13168728,smartcities7030060}. Complementing this, the regression analysis shows that pedestrian speed is influenced by multiple environmental factors, including pedestrian density, walkway width, slope, sidewalk quality, and temperature, and is reflected in robot operational factors such as robot speed and speed drops. These nuances emphasize that the environmental context may play a more important role in pedestrian behavior than the physical characteristics alone.

This study also reveals some underexplored mechanisms that can potentially shape the walking behavior. For instance, sidewalk irregularities and unevenness in this study emerged as a possible reason behind non-linear paths under crowded conditions. Prior studies \citep{silva2025engineering,buildings14061512} noted that uniform, high-quality surfaces improve comfort and reduce tripping hazards, but their potential to influence maneuvering behavior remains largely unexamined. In addition, sidewalks with greater effective width were, somewhat counterintuitively, associated with lower walking speeds. A plausible explanation is that wider spaces encourage relaxed pacing and lateral dispersion, particularly in outdoor, low-stakes settings. These findings warrant further investigation.

Overall, this work explores the practical potential of sidewalk delivery robots for collecting walkability-related features. Unlike traditional audits or fixed-position sensors, robots provide continuous, high-resolution measurements across entire routes, allowing features to be monitored over time. Moreover, robot trip data, such as speed and speed drops, can serve as reliable proxies for pedestrian movement, extending the scope of analysis beyond what conventional methods can capture. 
These data provide new opportunities for urban planners and policymakers to identify local bottlenecks, prioritize sidewalk maintenance, and assess the impact of interventions. For logistics operators, they offer practical information about the navigability and reliability of delivery routes under varying conditions. Importantly, realizing these benefits will require cooperation between cities and private operators to enable data sharing and to establish governance frameworks that balance commercial interests with public value. More broadly, robot-based sensing may help address the lack of granular data in walkability research, offering a way to build systematic, scalable, and inclusive assessments of pedestrian environments. 
 
However, this study presents some limitations. The experiments were performed on the KTH campus, which is a somewhat special environment. Data accuracy was partly constrained by the sensing and algorithmic capabilities of the robot, and the dataset size was limited (101 trips and 900 segment records). While these factors suggest caution in generalizing the results, the experiments provide valuable exploratory insights within the studied context, offering a foundation for further methodological and empirical extensions. 

Another limitation is that potential conflicts between sidewalk delivery robots and pedestrians were not explicitly addressed. Prior research indicates that these robots can be perceived as intrusive or unsafe and may cause conflicts at crosswalks, intersections, or narrow segments without clear lane delineations \citep{GEHRKE2023100789}.  Such interactions could affect both pedestrian and robot behavior, potentially introducing bias into measurements. These considerations underscore the importance of interpreting the results with an awareness of the broader social and regulatory context in which sidewalk robots operate.

\section{Conclusion}
\label{Sec:Conclusion}

This study demonstrates how sidewalk delivery robots can serve as scalable, automated platforms for collecting detailed walkability-related data in real time. By extracting a comprehensive set of features characterizing sidewalk conditions, pedestrian dynamics, and robot trip behavior, we provide a practical framework for continuous, granular assessment of walking environments without relying on expensive fixed infrastructure. Compared to prior work, the proposed approach captures a comprehensive range of walkability-related factors and offers new insights into the dynamic interaction between pedestrians and sidewalk infrastructure.

Our results show that pedestrian movement is closely tied to sidewalk characteristics such as density, width, and surface quality. Wider and smoother sidewalks tend to support steadier movement, but they can also encourage slower, more meandering walking in relaxed settings. These patterns confirm well-known relationships like the density–speed trade-off, while also refining them with context-specific insights from outdoor environments. Finally, the strong link between robot and pedestrian speeds suggests that delivery robots could serve as practical stand-ins for human movement, offering a new way to monitor and evaluate walkability. 

This study is limited to a campus setting with a modest dataset, and future work should test the approach in more varied urban contexts. Further research could also examine robot–pedestrian conflicts at crosswalks and narrow segments, as well as expand sensing capabilities to capture additional comfort and safety attributes such as pavement type, surface temperature, or slip resistance. Finally, exploring whether robot-based signals (e.g., vertical acceleration) reflect the (dis)comfort of users with strollers, wheelchairs, or trolleys may help develop more inclusive assessments of walkability. These directions could point toward a richer and more practical role for sidewalk robots in supporting walkable, human-centered cities.

\bibliographystyle{apalike}
\bibliography{Manuscript.bbl}

\end{document}